\documentclass[11pt]{article}

\usepackage[preprint]{acl}

\usepackage{times}
\usepackage{latexsym}

\usepackage[T1]{fontenc}

\usepackage[utf8]{inputenc}

\usepackage{microtype}

\usepackage{inconsolata}

\usepackage{graphicx}

\usepackage{amsmath}
\usepackage{enumitem}
\usepackage{booktabs}

\usepackage{soul}

\title{
MTDiag: A Multi-Turn Diagnostic Dataset Towards Clinically \\ Meaningful LLM Evaluation}

\author{
  \textbf{Pia Chouayfati\textsuperscript{1}},
  \textbf{Alexander M. Fichtl\textsuperscript{1}},
  \textbf{Miriam Anschütz\textsuperscript{1}},
  \\
  \textbf{George Doumat\textsuperscript{2}},
  \textbf{Georg Groh\textsuperscript{1}}
    \\[0.5em]  
  \textsuperscript{1}Technical University of Munich
  \textsuperscript{2}Department of Internal Medicine, UT Southwestern
    }

\usepackage[font=small]{caption}
\usepackage{xspace}
\newcommand{\pv}{{\small\texttt{PatientVector}}\xspace} 
\newcommand{\cui}{{\small\texttt{CUI}}\xspace} 
\newcommand{\cuis}{{\small\texttt{CUIs}}\xspace}

\begin{document}
\maketitle
\begin{abstract}

Clinical diagnosis is fundamentally interactive and incremental, yet the dominant paradigm for evaluating Large Language Models (LLMs) in medicine remains static QA benchmarks or template-based dialogues. These benchmarks say little about whether a model can serve as a diagnostic agent in a dynamic clinical encounter, with LLMs showing significant accuracy and reliability degradation in multi-turn settings.
To address this issue, we present MTDiag, a large multi-turn diagnostic dialogue dataset constructed from three heterogeneous sources: DDXPlus, MIMIC-IV, and published case reports (AJCR), covering common ED presentations as well as long-tail rare and atypical conditions. All cases are normalized into a canonical \pv schema anchored in the most comprehensive and widely-adopted medical knowledge bases (UMLS concept identifiers, with ICD-10 diagnosis codes). We release the \pv schema, a UserLM-8B-based utterance-generation pipeline, and the physician-validated dataset that converts structured clinical evidence into natural-language utterances. Importantly, we introduce and motivate clinical knowledge-grounded metrics for evaluating LLMs as diagnostic agents, beyond diagnostic accuracy, for the task of multi-turn differential diagnosis.

\end{abstract}

\section{Introduction}
\vspace{-5pt}
Large language models (LLMs) are increasingly seen as viable tools to enhance healthcare by assisting physicians in their daily tasks \cite{vladika2026investigating}, but their suitability for direct patient-facing settings requires extensive testing and evaluation.
The dominant approach for evaluating LLMs in medicine has long been the static multiple-choice examination or question-answering (QA) set-up \cite{jin2019pubmedqa, jin2021disease, nentidis2025bioasq}. This approach has driven measurable progress, with frontier models routinely achieving near-perfect scores on medical licensing benchmarks. However, a recent systematic review of 39 clinical LLM benchmarks highlights a severe knowledge-practice gap: while models achieve up to 90\% accuracy on knowledge-based questions, performance drops to roughly 45\% on practice-based diagnostic tasks \cite{gong2025knowledge}. This degradation occurs because clinical diagnosis is fundamentally an interactive and incremental process of information extraction, prioritization, and synthesis under uncertainty. Existing benchmarks present models with complete, pre-structured, and often self-contained information. In reality, a physician must actively elicit information from a patient who may forget key symptoms, use imprecise language, or omit critical context. High performance on a vignette benchmark is therefore not predictive of real-world diagnostic competence \cite{alaa2025position}.


Recent efforts toward conversational diagnostic AI \cite{tu2025towards, fan-etal-2025-ai, johri2024craft} confirm this, finding clear accuracy and reliability degradation in interactive settings, while millions of users already seek medical guidance from general commercial LLMs. Alarmingly, a recent qualitative analysis by the medical community of patient conversations with commercial LLMs has shown that LLMs provide unsafe answers to patient-posed medical questions \cite{draelos2026large}.

\begin{figure}[t]
  \includegraphics[width=\columnwidth]{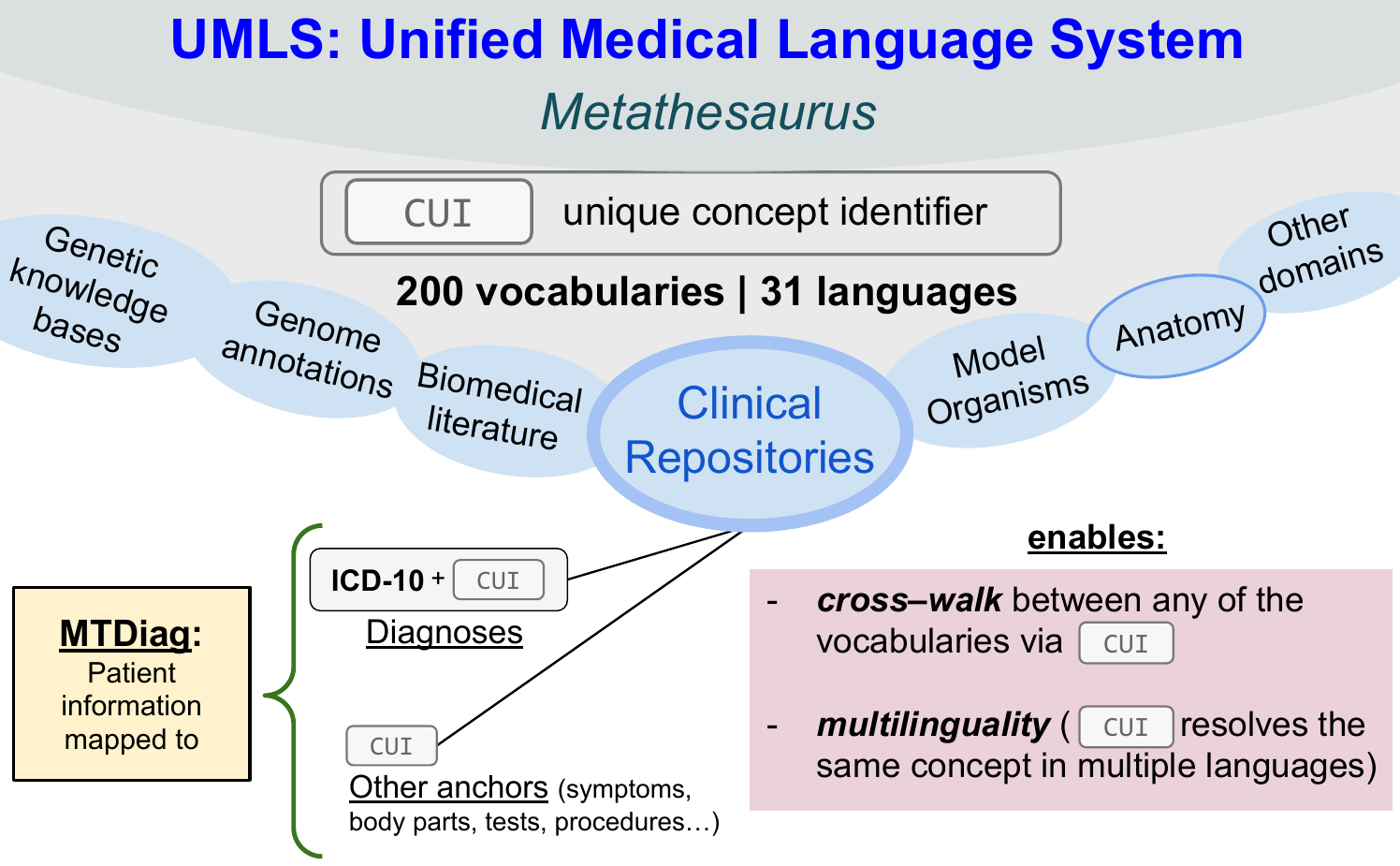}
\caption{The UMLS (Unified Medical Language System) \cite{bodenreider2004unified} is a metathesaurus of international biological and medical knowledge bases, ontologies and vocabularies built for healthcare and medical system interoperability. It indexes unique concepts by \cui (unique concept identifier), and instantiates a semantic network (Fig. \ref{fig:umls_subset}) which links a single concept across all participating source vocabularies.}
\label{fig:umls_main}
\end{figure}

To address this gap, we present \textbf{MTDiag}, a large multi-turn diagnostic dialogue dataset constructed from three heterogeneous real and realistically simulated clinical sources and anchored in the UMLS (Fig. \ref{fig:umls_main}), which enables clinical reasoning evaluation beyond diagnostic accuracy. Our primary contributions are:

\begin{itemize}[noitemsep,topsep=0pt]
    \item[\textbf{(1)}] \textbf{MTDiag Dataset}: A multi-turn diagnostic dataset normalizing data from DDXPlus \cite{fansi2022ddxplus}, MIMIC-IV \cite{johnson2023mimic}, and a curated diagnostic AJCR case compilation \cite{hirosawa2024comparative} into a canonical \pv schema (Fig. \ref{fig:PVMAIN}), where diagnoses are mapped to ICD-10\footnote{ICD-10 (Figure \ref{fig:icd10}): 10th revision of the International Classification of Diseases, maintained by the (WHO) World Health Organization}, and symptoms are mapped to UMLS (Fig. \ref{fig:umls_main}) \cui identifiers. 
    \item[\textbf{(2)}] \textbf{Utterance Generation Pipeline}: A UserLM-8B \cite{naous2025flipping} based framework that converts structured clinical evidence into natural, patient-like utterances. It allows case-specific generation of utterances and presentations, providing a set of canonical utterances per \pv.

    \item[\textbf{(3)}] \textbf{Patient Orchestrator}: A MedGemma \cite{sellergren2025medgemma} based runtime agent that selects contextually appropriate responses from the pre-generated utterance pool. This "dialogue runner" simulates a dialogue between a patient and an examiner via the \pv. 

\end{itemize}

\begin{figure}[t]
  \includegraphics[width=\columnwidth]{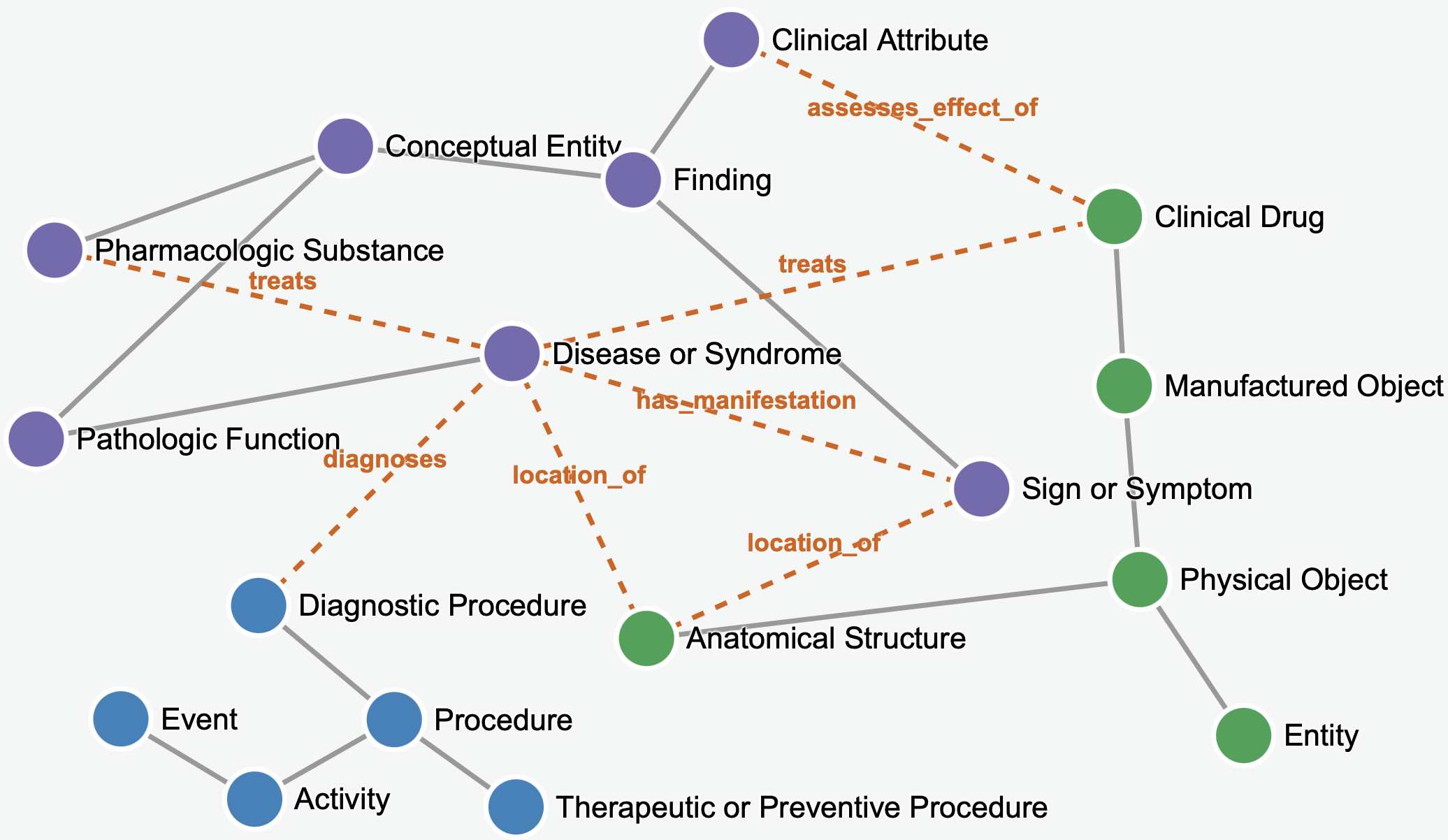}
  \captionsetup{singlelinecheck=off}
\caption{Clinically-relevant subset of the UMLS Semantic Network, showing Semantic Type
nodes and typed relation edges. Nodes are
color-coded by semantic category: 
\newline\textcolor[HTML]{3182bd}{$\bullet$}~\textbf{Events}
(actions occurring in time), e.g.\ Procedure
\newline\textcolor[HTML]{31a354}{$\bullet$}~\textbf{Entities} (physical matter and anatomy) e.g.\ Clinical Drug
\newline\textcolor[HTML]{756bb1}{$\bullet$}~\textbf{Conceptual Entities} (abstract classifications and states), e.g.\ Disease or Syndrome \newline Solid edges denote hierarchical \textit{is a} relations; dashed edges denote typed associative relations.}
\label{fig:umls_subset}
\end{figure}
\vspace{-5pt}
\section{Background and Related Work}
\vspace{-4pt}
MTDiag is designed to enable clinically-grounded simulation of the "diagnostic encounter," which we define as:  a \textbf{patient} willingly seeking a \textbf{diagnosis} for a \textbf{"chief complaint" (CC)}, via a \textbf{dialogue} (examination) with a \textbf{medical practitioner} (examiner).  The examiner's task is inherently one of
\textbf{differential diagnosis}: the iterative (and implicit) process of generating, ranking, and progressively narrowing a set of candidate diagnoses by eliciting and weighing clinical evidence turn by turn, until a most probable diagnosis can be committed to \cite{fansi2022ddxplus, tu2025towards}. MTDiag's structure thus indirectly models this process from the examiner side: the ground-truth diagnoses, symptom and history anchors, and dialogue logs serve to evaluate whether a model is correctly reasoning differentially: asking the right questions, in the right order, for the right reasons.
In the following sections, we establish why this encounter is structurally different from what current benchmarks capture and survey the prior work that motivates and contextualizes the dataset design.
\vspace{-4pt}
\subsection{The Diagnostic Encounter as a Dialogue Modeling Task} 
\label{sec:encounter}
\vspace{-2pt}

Physicians gather information from patients who may be vague, forgetful, resistant, or unaware of what is clinically relevant. The history of present illness (HPI), a log of the patient's medical past, is never handed over; it is negotiated across a conversation \citep{redelmeier2001problems}, with patients routinely exhibiting systematic failure modes in reporting their history, including errors in comprehension, recall, and expression.  Vague patient presentations reduce the discriminative power of the symptom and lower the probability of a correct diagnosis \cite{sonnenberg2002translating}. Patients will also actively resist or renegotiate diagnostic framings after hearing a physician's assessment \citep{ijas2010patient}, a dynamic that static vignettes cannot model. Furthermore, clinical reasoning, whether human or artificial, is vulnerable to cognitive biases that compound these challenges \citep{vally2023errors}.
In realistic scenarios, LLMs exhibit inconsistent risk-stratification \cite{heston2024chatgpt}, fail to follow diagnostic guidelines during incremental information gathering \cite{hager2024evaluation}, and alter factual clinical outputs based on patient linguistic markers \cite{kearney2025language, zhou2025unveiling}. As a medical direct-to-consumer offering, ChatGPT Health undertriaged up to 52\% of emergencies and shifted recommendations when patients minimized their symptoms \cite{ramaswamy2026chatgpt}. The questions MTDiag answers are therefore not only \textit{does the model arrive at the right diagnosis?} but more importantly: \textit{can the model conduct a diagnostic conversation well enough to reach the right diagnosis?}
\vspace{-7pt}

\paragraph{Static Medical QA Benchmarks}
The standard for assessing the "clinical aptitude" of LLMs (and making the case for them as "medical assistants") has mostly relied on increasingly strong results on static QA benchmarks such as PubMedQA \cite{jin2019pubmedqa}, MedQA \cite{jin2021disease}, and MedMCQA \cite{pal2022medmcqa}, which aggregate questions from medical licensing exams and literature. Questions on medical exams are self-contained: with adequate knowledge, everything required for the answer is in the question formulation. This is far from real patient records:
Because they test recall under ideal conditions, correctly
answering a question when presented with a QA vignette does not predict correct performance on similar real-world cases \cite{alaa2025position, gong2025knowledge}. The validity of these benchmarks is further undermined by data contamination and leakage \cite{balloccu-etal-2024-leak, xu2024benchmarking}, where test items appearing in pretraining corpora inflate reported scores.
\vspace{-7pt}
\paragraph{Conversational Diagnostic AI}
Recent systems have begun probing LLMs in interactive clinical settings.
AMIE \cite{tu2025towards} and CRAFT-MD \cite{johri2024craft} demonstrate that models capable of near-perfect performance on static benchmarks degrade substantially when required to elicit information across a multi-turn encounter. AI Hospital \cite{fan-etal-2025-ai} and the JAMA multi-agent framework \cite{sangwon2025evaluating} further surface reliability and consistency failures in interactive settings. \citet{arias2025automatic} propose automatic evaluation of healthcare LLMs beyond question-answering, and implement a range of sub-tasks, motivating the need for dialogue-level assessment. 
\vspace{-7pt}
\paragraph{Medical Dialogue Datasets}
Several medical dialogue datasets have been introduced to capture multi-turn dynamics, most recently surveyed by \citep{gong2025knowledge}. We describe the datasets we considered and their suitability in Section~\ref{sec:selecteddatasets}. 

\vspace{-5pt}
\section{Ontological Anchoring and Resource Choices}
\label{sec:ontological_anchoring_resource_choices}
\vspace{-5pt}
In this section, we discuss the design decisions behind the construction of MTDiag, which were directly informed by practicing medical professionals and by surveying well-established clinical behavioral literature, taking into account a space of logical and practical errors that can occur in a multi-turn differential diagnosis setting. We then introduce and motivate some of the metrics that ontological anchoring enables.

\vspace{-5pt}

\paragraph{Minimum Anchors} We first define three necessary elements (which we refer to as our "Minimum Anchors") for the conduction and evaluation of a diagnostic dialogue:

\begin{figure}[t]
  \includegraphics[width=\columnwidth]{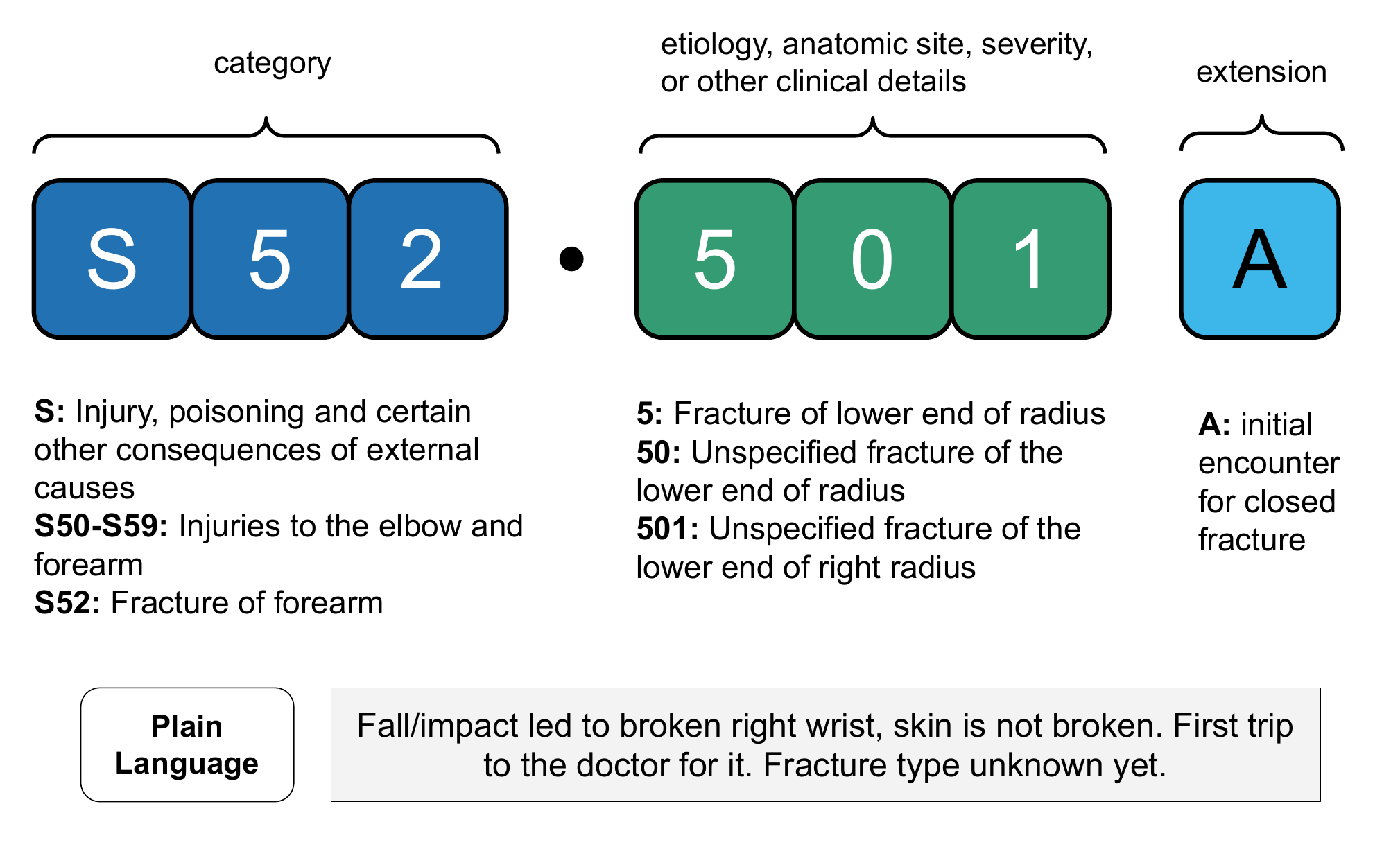}
  \vspace{-5mm}
  \setlength{\belowcaptionskip}{-15pt}
\caption{ICD-10-CM code structure illustrated with \texttt{S52.501A}
(unspecified fracture of the lower end of the right radius, initial
encounter for closed fracture). Each position encodes a specific
clinical dimension: category, anatomic site/etiology, severity, and
encounter extension. The WHO ICD-10 equivalent without the CM extension would be \texttt{S52.5}.}
  \label{fig:icd10}
\end{figure}

\begin{figure*}[t]
  \includegraphics[width=0.99\textwidth]{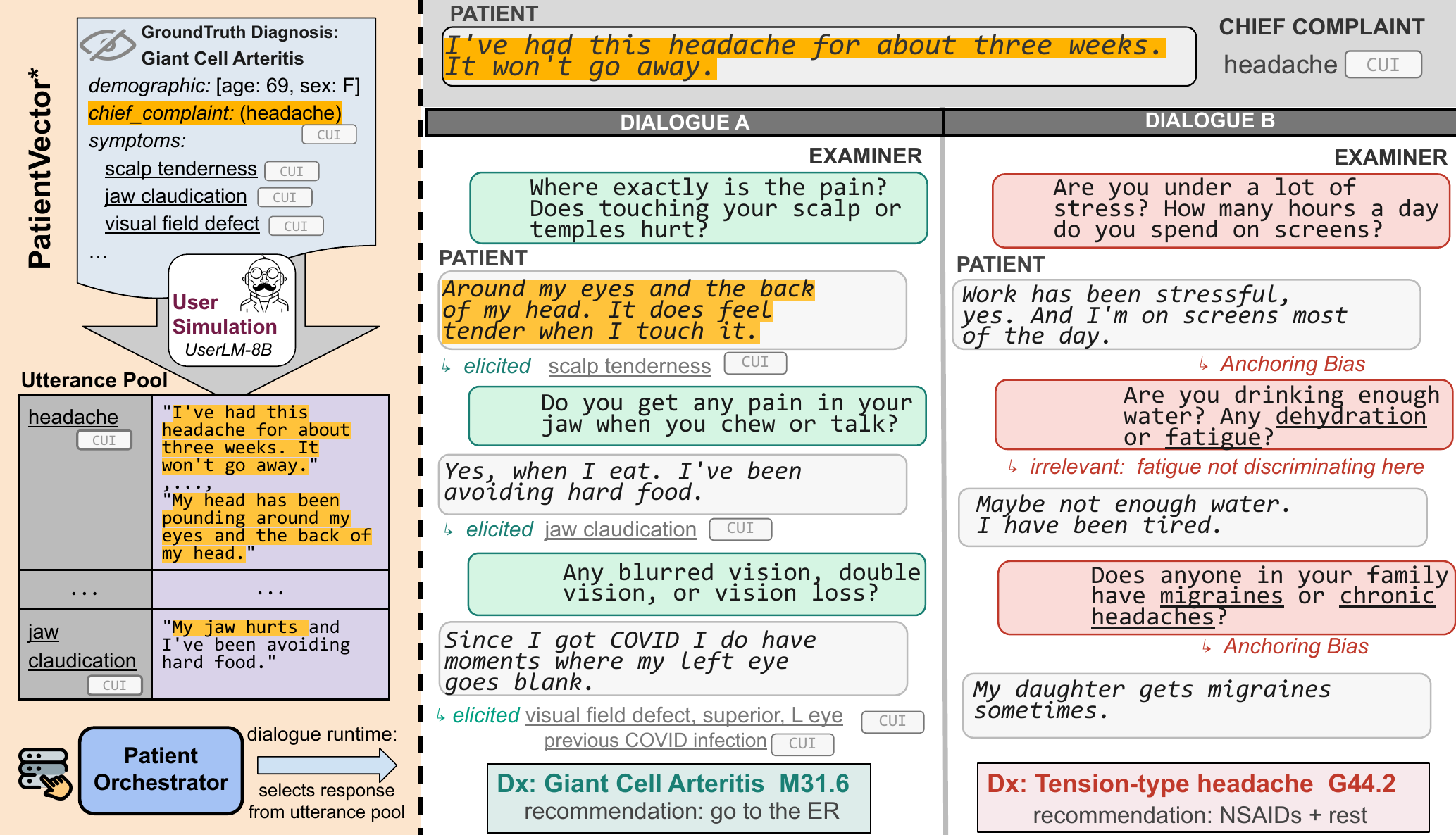}
    \centering
  \caption{(Condensed) example of ontology-enabled dialogue evaluation - MTDiag AJCR case. Dx: Giant Cell Arteritis (GCA, M31.6) \cite{szydelko2022arteritic}. \textbf{Left} (of dashed line): the \pv* for this case, comprising the ground-truth diagnosis (masked from examiner) and UMLS \cui-mapped symptoms (subset only). Utterances are pre-generated offline by UserLM-8B for each element of the \pv, constituting the \cui-mapped utterance pool. At dialogue runtime, a Patient Orchestrator LLM answers examiner questions from the utterance pool.
\textbf{Right:} Dialogue runtime: Two dialogues conducted from the same chief complaint utterance.
\textbf{Dialogue~A}: examiner probes for scalp
tenderness, jaw claudication, and visual field loss and reaches the correct diagnosis.
\textbf{Dialogue~B}: examiner anchors prematurely on migraine/tension headache, asks only about stress, screen time, hydration, and family history, and diagnoses tension-type headache. A \textbf{Harm~Tier~2} error (delay of care) occurs. *simplified}
\label{fig:worked_example}
\end{figure*}

\begin{itemize}[noitemsep,topsep=0pt]
    \item The \textbf{chief complaint (CC)}: the patient's primary reason to seek care, according to "them." This serves as the opening utterance of the dialogue (turn 0).
    \item \textbf{Symptoms}: the patient's other symptoms.
\item The \textbf{primary diagnosis}: the ground truth principal finding of the case, as ICD-10 code (Fig. \ref{fig:icd10}). An ICD-10 code match constitutes the minimum evaluation target: did the model reach the right diagnosis?
\end{itemize}

\vspace{-5pt}

\paragraph{Dialogue Trace} To illustrate why this minimum evaluation target of Diagnosis Accuracy ($\mathrm{Diag_{acc}}$) fails to completely capture clinical competence, we trace two example dialogues, shown in Figure \ref{fig:worked_example}.
Both LLM examiners receive the same opening patient utterance: \textit{"I've had this headache for about three weeks. It won't go away."} The ground-truth diagnosis is Giant Cell Arteritis (GCA, \texttt{M31.6}), a serious inflammatory disease of the arteries. In \textbf{Dialogue A}, the examiner probes for scalp tenderness, jaw claudication, and visual changes. After the patient confirms these symptoms, the examiner correctly diagnoses GCA. In \textbf{Dialogue B}, the examiner immediately anchors on a migraine hypothesis, and asks only questions that reinforce that initial hypothesis (stress, screen time, hydration, and family history), ultimately diagnosing a Tension-Type Headache (\texttt{G44.2}) and completely missing GCA. 
\vspace{-5pt}
\paragraph{Dialogue Metrics} In addition to a classification error, which is trivially detected by ICD-10 mismatch, \textbf{Dialogue B} has two further issues. The examiner displays \textit{\textbf{anchoring bias}} (a line of questioning that serves only to confirm an initial hypothesis rather than distinguish it from alternatives) towards migraine/tension-type headache. \cui anchoring makes this detectable by mapping the examiner questions to the symptom cluster for migraine, with no questions mapped to the cluster for GCA. Another more severe issue with Dialogue B is that it ultimately leads to a \textit{\textbf{Harm Tier 2}} error (delay of care), where the incorrect diagnosis leads to delay in seeking treatment for GCA, which can cause irreversible bilateral blindness if left untreated.
A full discussion of the \textbf{Harm Index} and formal definitions of other ontologically-enabled diagnostic dialogue metrics follows in Section \ref{sec:discussion}.
\vspace{-5pt}
\subsection{Patient-Chatbot Realism}
\label{sec:patientchatbotrealism}
\vspace{-5pt}
Assessing LLMs in user-facing diagnosis means the "patient" can realistically communicate their symptoms via a chat interface, as an alternative to seeking medical care. This rules out patients who are unconscious, intubated, or in agonizing pain. These and more considerations guide an extensive pre-processing of the datasets detailed in Section \ref{sec:preprocessing}.
Similarly, the need to realistically simulate a user interacting with an LLM rather than speaking to a doctor led us to employ UserLM-8B \cite{naous2025flipping} for utterance generation, detailed in Section \ref{sec:utterancegen}. Unlike standard instruction-tuned, "assistant-style" models, UserLM-8B is trained on user-chats with ChatGPT and optimized for "user-simulation," allowing colloquial, incomplete expression with the natural hedging and imprecision characteristic of lay speech.
\vspace{-5pt}
\subsection{Source Datasets}
\label{sec:selecteddatasets}
\vspace{-3pt}
Three sources were ultimately selected for MTDiag (full discussion in Appendix \ref{app:unuseddatasets}) to cover complementary regions of the diagnostic space: a structured synthetic head, a real high-frequency ED cohort, and a long-tail of rare and atypical case reports.

\vspace{-5pt}
\paragraph{DDXPlus \citep{fansi2022ddxplus}} is a large-scale synthetic dataset for differential diagnosis comprising 1.3~million patient cases across 49~pathologies and 223~evidences (symptoms and antecedents). Each record includes a differential diagnosis list with associated probabilities and a ground-truth label, as well as associated symptoms for all covered diseases.
\textbf{MIMIC-IV \citep{johnson2023mimic}} is a large, de-identified electronic
health record database from the Beth Israel Deaconess Medical Center covering over 200,000 ED (Emergency Department) visits and 65,000 ICU (Intensive Care Unit) stays (2008--2019). It provides real, high-frequency emergency presentations grounded in verified ICD-10 discharge diagnoses, augmented
by community extensions supplying narrative HPI text, structured laboratory results, and radiology reports.
\textbf{AJCR Case Reports \citep{hirosawa2024comparative}} is a curated compilation of 392 adult case reports from the \textit{American Journal of Case Reports} (January 2022--March 2023), with clearly stated final diagnoses. Case reports are selected precisely because they document rare, atypical, or diagnostically challenging
presentations (the long tail of diagnostic space that DDXPlus and MIMIC-IV do not
cover). Malignancies, diverse infections, and vascular diseases account for over 60\%
of final diagnoses.

\begin{figure*}[t]
  \includegraphics[width=0.99\linewidth]{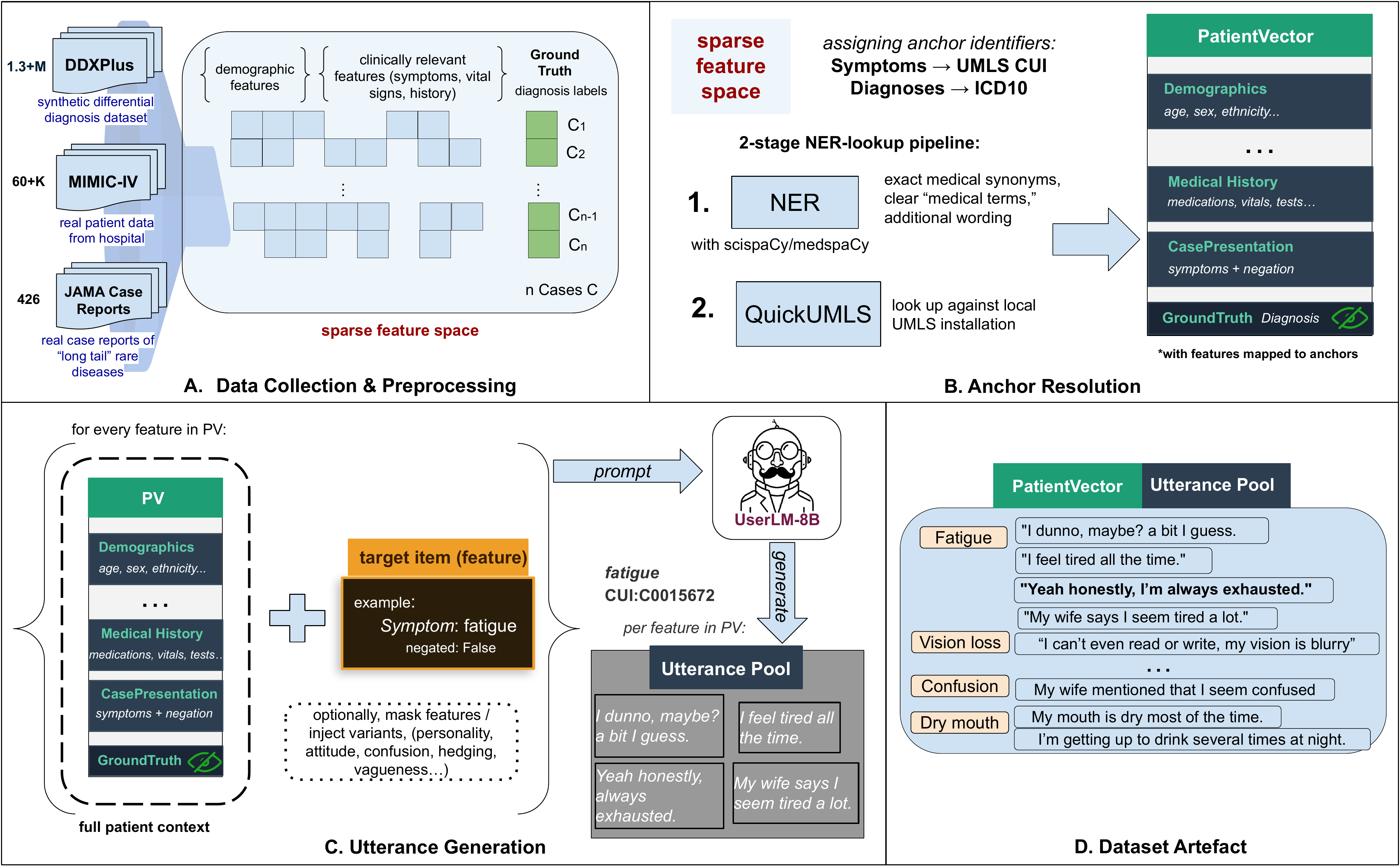} \hfill

\caption{Construction of MTDiag
\textbf{A. Data Collection \& Pre-Processing}: the three source datasets
(DDXPlus, MIMIC-IV, AJCR) and their heterogeneous sparse feature spaces,
preprocessed to satisfy the minimum anchor requirements and patient realism. \textbf{B. Anchor Resolution}: clinical strings from all three sources are segmented by \texttt{medspaCy} NER and matched against a local UMLS installation via \texttt{QuickUMLS} \cite{soldaini2016quickumls}, yielding a \cui and \texttt{mapping\_confidence} score per span. Diagnoses follow a separate ICD-10 crosswalk path. The pipeline runs fully locally against a fixed UMLS release, producing reproducible, verifiable mappings with no external API calls.
\textbf{C. Utterance generation: } 
For each item in \texttt{CasePresentation},
UserLM-8B is conditioned on \pv context and generates candidate utterances spanning expression styles (precise, colloquial, third-party); masking and persona parameters or noise injections  can be injected at generation time. Elements of the \pv context can be selectively masked to produce presentation variants. \textbf{D. Dataset Artefact: }
The \textbf{dataset proper} consists of the \pv, and the \textbf{dataset artefact}, is the result of utterance generation over the \pv. This artefact is structured, reproducible, and fully decoupled from the dialogue runtime described in Section~\ref{sec:orchestrator}.}
  \label{fig:dataset_artefacts}
\end{figure*}

\section{Methodology}
The construction of MTDiag is illustrated in Figure~\ref{fig:dataset_artefacts}. First, (Figure~\ref{fig:dataset_artefacts}\textbf{.A}; Section \ref{sec:preprocessing}) the three source datasets are collected and preprocessed. Then,  (Figure~\ref{fig:dataset_artefacts}\textbf{.B}; Section \ref{sec:anchor_pv}) the clinical evidence anchors (symptoms, diagnoses, medications...) are resolved to canonical ontological identifiers via a two-stage NER-to-UMLS pipeline, producing an annotated representation of each case. Each case is then normalized into the \pv unified schema, then (Figure~\ref{fig:dataset_artefacts}\textbf{.C}; Section \ref{sec:utterancegen}) the clinical evidence in the \pv is converted offline into natural-language patient utterances. 
\vspace{-5pt}
\subsection{Data Collection \& Preprocessing}
\label{sec:preprocessing}
\vspace{-5pt}
The three source datasets introduced in Section~\ref{sec:selecteddatasets} are structurally heterogeneous but were selected to satisfy the minimum anchor requirements defined in Section \ref{sec:ontological_anchoring_resource_choices}. \textbf{Chief complaint (CC)}, \textbf{symptoms} and \textbf{primary diagnosis} are present (or extractable), well-defined, and verified. However, the source datasets encode significantly more information than the minimum anchors. 
\textbf{Secondary diagnoses} are additional diagnoses recorded for the same case, capturing comorbidities and incidental findings. These are used to assess diagnostic completeness and to flag cases where the model's diagnosis, while not matching the primary diagnosis exactly, is clinically consistent with the documented picture and may match one or more of the secondary diagnoses. Medical history antecedents, narrative HPI, medications, lab tests, demographic and 
socioeconomic features, and more are also present with varying sparsities. The \pv (Fig. \ref{fig:PVMAIN}) schema is the canonical form normalizing all three sources, discussed in Section~\ref{sec:anchor_pv}.

Below, we document the key curation and preprocessing decisions for each source, with full pipeline details in Appendix~\ref{app:sourcedatasets}.

\begin{table}[t]
  \centering
  \small
\begin{tabular}{llr}
  \hline
  \textbf{Source} & \textbf{Type} & \textbf{Scale (cases)} \\
  \hline
  DDXPlus   & Synthetic & 1.3M \\
  MIMIC-IV  & Real EHR  & 68K     \\
  AJCR      & Case reports & 415    \\
  \hline
\end{tabular}
\setlength{\belowcaptionskip}{-15pt}
  \caption{Data sources used in MTDiag after filtering and deduplication (detailed in Appendix~\ref{app:mimic_filtering}).}
  \label{tab:datasource_summary}
\end{table}

\subsubsection{DDXPlus} 
Synthesized from a proprietary Automatic Disease Diagnosis (ADD) system, DDXPlus encodes structured clinical evidence (symptoms, antecedents) under a closed-world assumption, with each case carrying a differential diagnosis list and ground-truth label. Its value for MTDiag lies in its differential trajectories and structured evidence instantiation. One preprocessing choice worth mentioning here: each patient's \texttt{INITIAL\_EVIDENCE} field (a randomly selected binary evidence with no inherent clinical salience in the original ADD setup) is retained as the CC for consistency.

\subsubsection{MIMIC-IV}
\label{sec:mimiciv}
 \vspace{-3pt}
Unlike typical uses of MIMIC for longitudinal time-series applications, our use case requires coherent narrative patient presentations.  We draw on four official modules (Hosp, ED, Note) and two community extensions (BHC hospital course narratives; CDM curated abdominal pathology cases).
\vspace{-3pt}
\paragraph{Population selection}
The target population is patients who walk into an ED of their own volition and could plausibly interact with an LLM instead.  From 425,087~raw ED visits, we sequentially exclude ambulance arrivals, visits with pain score above~8/10 or ESI acuity level~1, cases with no primary diagnosis (etc.), and deal with free-text entries for some numeric values as "extra context" which we preserve. The preprocessing details are in Appendix~\ref{app:mimic_filtering}.
\vspace{-3pt}
\paragraph{Two-track design}
After filtering, the cohort partitions into two clinically distinct populations.
\textbf{Track~1 (ED Discharge):} The patient is admitted into the Emergency Department, then discharged. Only ED data is used (chief complaint, vitals, questionnaire); ground truth is the ED diagnosis.
\newline
\textbf{Track~2 (Hospital Admission):} The patient is admitted into the ER, evaluated, and then admitted to the hospital for further work-up. This augments triage (ED) data with the History of Present Illness and, where available, granular structured clinical data from the BHC and CDM extensions. The CDM  extension alone contributes pre-extracted HPI narratives, physical examination findings, 138,788 laboratory results across 480 unique tests, 4,403 microbiology results, and 5,959 radiology reports (CT, X-ray, ultrasound, MRI) with diagnostic conclusions explicitly stripped from radiology findings and cases where the pathology name appears in the HPI excluded, making these narratives genuinely suitable for diagnostic simulation without leakage.

\vspace{-7pt}

\paragraph{Deduplication}
A hierarchical strategy retains one visit per patient (to prevent "frequent flyers" from saturating the baseline), prioritized by data richness (CDM + hospital course $>$ CDM only $>$ hospital course only $>$ earliest visit). The final subset comprises \textbf{68,346 unique patients}: 49,440 ED-discharged (Track~1) and 18,906 admitted (Track~2).  

\vspace{-5pt}
\paragraph{Zero Data Retention} As MIMIC-IV consists of de-identified yet real patient data, its use is subject to PhysioNet guidelines and processing is only permitted under ZDR (Zero Data Retention). As such, all processing runs on the GWDG KISSKI secure HPC. 

\vspace{-5pt}
\subsubsection{AJCR Case Reports}
\vspace{-4pt}
We developed a general-purpose scraping and parsing pipeline for AJCR articles: given a DOI, it retrieves full-text HTML and extracts structured metadata, abstracts, section-split body text, figures, and tables into a per-case JSON schema, with automatic detection and splitting of case series. We apply it here to the Hirosawa et al.~\cite{hirosawa2024comparative} compilation precisely because it provides pre-curated adult diagnostic cases with verified final diagnoses, yielding 415 cases from 392 reports (see Table \ref{tab:datasource_summary}). The pipeline is applicable to any AJCR DOI.
\vspace{-5pt}
\subsection{Normalization and Anchor Resolution}
\label{sec:anchor_pv}
\vspace{-3pt}
\subsubsection{\pv Unified Case Schema}
\label{sec:patientvector}
\vspace{-3pt}
To unify the three heterogeneous sources into a single structure, we define the \pv (Figure \ref{fig:PVMAIN}), a canonical per-case data schema that all sources are normalized into prior to dialogue generation. The schema comprises four top-level blocks:

\begin{figure*}[t]
  \includegraphics[width=0.99\linewidth]{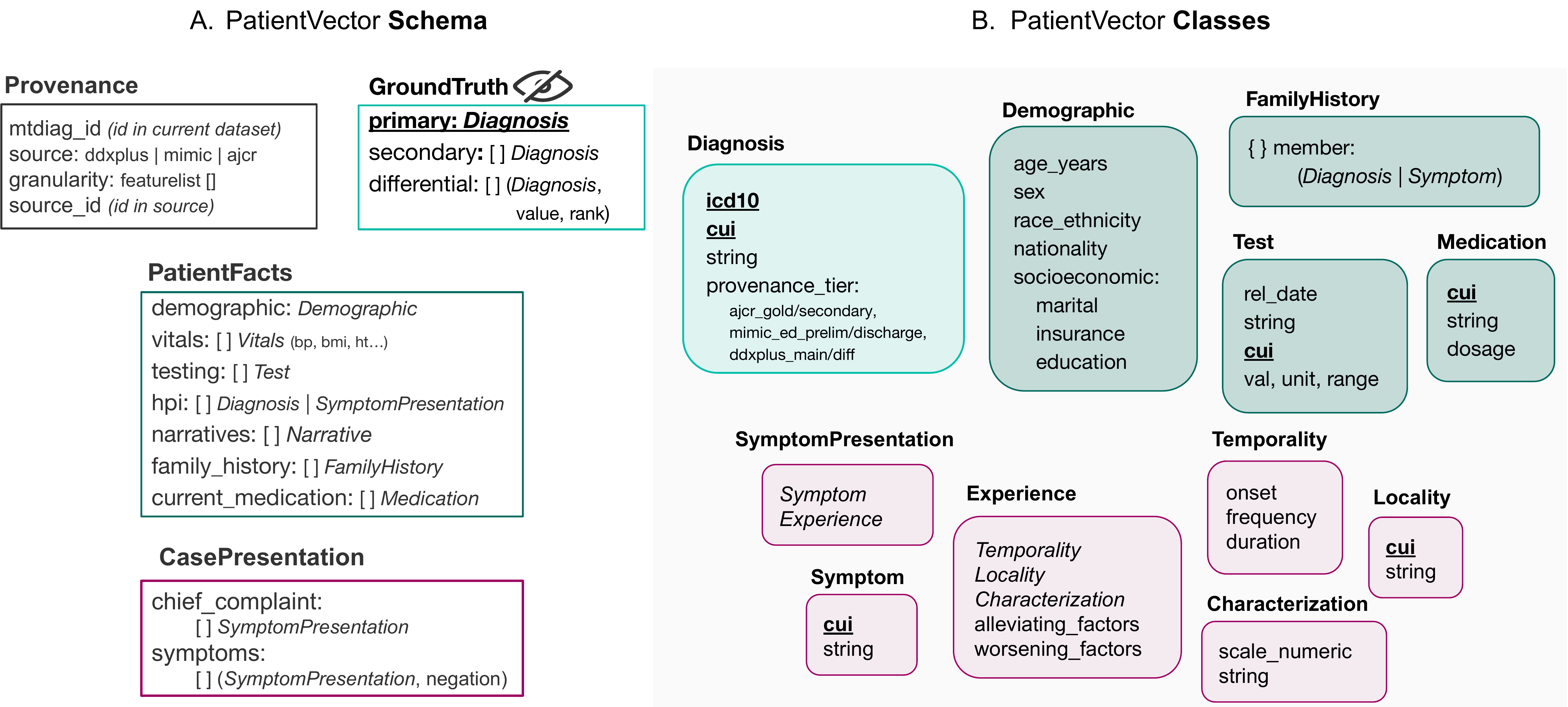} \hfill

\caption{\pv schema. \textit{Left}: the four top-level
blocks: Provenance, GroundTruth (masked), PatientFacts, and CasePresentation, with their fields. \textit{Right}: the full class hierarchy, showing how SymptomPresentation decomposes into Symptom (\cui + string) and Experience (Temporality, Locality, Characterization), and how Diagnosis carries ICD-10 code, \cui, and \texttt{provenance\_tier}. All fields are serialized to JSON; sparsity varies by source. *ICD-10 is also indexed in the UMLS, so every ICD-10 code resolves to a \cui.}
  \label{fig:PVMAIN}
\end{figure*}

\textbf{Provenance} records the origin of each case: a globally unique \texttt{mtdiag\_id}, the source dataset (\texttt{ddxplus}, \texttt{mimic}, or \texttt{ajcr}), the native identifier in that source, and pipeline metadata.  This block enables stratified evaluation and ablation by source.

\textbf{GroundTruth} is populated at ingestion and never exposed to the dialogue generation pipeline or the examiner.  It contains the primary \texttt{Diagnosis} (ICD-10 code, UMLS \cui, and human-readable label), a list of secondary \texttt{Diagnosis} entries, and, where available, a ranked differential diagnosis list.  Each diagnosis carries a \texttt{provenance\_tier} field recording the reliability of the ground-truth label (distinguishing, for instance, a verified hospital discharge diagnosis (\texttt{mimic\_discharge}) from a preliminary ED assessment (\texttt{mimic\_ed\_prelim})) (additional details in Appendix \ref{app:patientvector}).

\textbf{PatientFacts} holds all patient information available beyond the presenting complaint: demographics (e.g., age, sex, race/ethnicity, nationality, and socioeconomic proxies like insurance, marital status), vitals (blood pressure, BMI, height), test results (\cui-mapped, with value, unit, reference range, and relative date), narrative text (HPI and free-text notes), family history, and current medications (\cui, name, and dosage).  The sparsity varies by source: MIMIC-IV Track~2 cases with records in the CDM and/or BHC extensions provide data for most fields, while DDXPlus cases carry only some demographics and binary/categorical evidence.

\textbf{CasePresentation} is the snapshot of the patient as they initiate the dialogue. It contains the chief complaint: one or more \texttt{SymptomPresentation} entries forming the opening utterance and the full symptom list.  Each \texttt{SymptomPresentation} pairs a \texttt{Symptom} (\cui and string) with an \texttt{Experience} capturing temporality (onset, frequency, duration), locality (\cui-mapped body region), characterization (severity scale and/or free-text descriptor), and aggravating/relieving factors. 
The full field-level specification of all classes is provided in Appendix~\ref{app:patientvector}.
\vspace{-5pt}
\subsubsection{Anchor Resolution}
\vspace{-5pt}
The core design feature of MTDiag is that all clinical anchors (and additional patient facts) are mapped to established canonical identifiers: the chief complaint, symptoms, and other details where possible (lab tests, medications...) are mapped to UMLS \cuis (Fig. \ref{fig:umls_subset}); diagnoses are mapped to ICD-10 codes (Fig.~\ref{fig:icd10}).
A UMLS \cui situates a concept within a graph of over 200 integrated clinical vocabularies. Rather than asking whether two symptom strings match, one can ask how far apart two concepts are in the UMLS graph (Fig.~\ref{fig:umls_subset}), whether one subsumes the other, or whether they share a common parent, resulting in a graded, clinically meaningful similarity measure. It also enables crosswalk between vocabularies: a symptom expressed in lay language and its formal medical term (in any of the included vocabularies) resolve to the same \cui. Mapping the heterogeneous and sparse feature spaces of our source datasets to \cuis proved non-trivial; both LLM-based and UMLS API string search-based resolution proved unsuitable for large-scale annotation; we discuss these failure modes in Appendix~\ref{app:anchorresolution}.
Our approach (Figure \ref{fig:dataset_artefacts}.B) relies on a 2-stage NER-based pipeline. In the first stage, clinical named entity recognition is performed using \texttt{medspaCy} \cite{medspacy}, a spaCy-based framework for clinical text processing, which segments input text and identifies candidate medical concept spans.  In the second stage, each candidate span is passed to \texttt{QuickUMLS}~\cite{soldaini2016quickumls}, a fast approximate string matching engine built over a local UMLS installation, which returns the best-matching \cui along with a similarity score retained as the \texttt{mapping\_confidence}.  All additional non-anchor patient information is mapped back to either \cui identifiers, tightly structured enumeration options, or numeric scale representations.
\vspace{-5pt}
\subsection{Utterance Generation}
\label{sec:utterancegen}
\vspace{-3pt}
The \pv schema is designed to support a spectrum of patient variability. To convert the structured symptom representations into natural language, we use UserLM-8B \cite{naous2025flipping}, a model trained to produce human-like \textit{user} utterances (Figure \ref{fig:dataset_artefacts}.C).
 We provide a set of canonical utterances per \pv, which can be edited and re-generated via this pipeline.  All utterances are pre-generated, stored, and fully decoupled from live dialogue. The patient's "factual reality" is fixed at dataset creation time, rather than simulated during a dialogue.
This "separation of concerns" ensures reproducibility, prevents the patient simulator from hallucinating new symptoms, and allows human review of the utterance pool. 
\vspace{-5pt}
\subsection{Dialogue Runtime}
\label{sec:orchestrator}
\vspace{-3pt}
At dialogue runtime, two agents interact (illustrated in Fig.~\ref{fig:patientorchestrator} in the Appendix).  The \textbf{Patient Orchestrator}, powered by MedGemma \cite{sellergren2025medgemma}, receives the examiner's most recent question and selects which pre-generated utterance or combination of utterances to respond with. MedGemma was chosen due to being the only medical-specialized and instruction-tuned model available via GWDG ChatAI, our ZDR-compliant inference provider, but it is possible to use any assistant LLM as Patient Orchestrator. The orchestrator draws exclusively from the pre-generated utterance pool, ensuring responses remain grounded in case facts. This design deliberately separates \textit{content} (what the patient says, grounded and \cui-anchored) from \textit{selection logic} (which utterance to surface given the conversation), as a middle ground between the rigidity of rule-based response selection and the hallucination risk of free-form patient simulation. The \textbf{Examiner} is the model under evaluation.  It receives only the chief complaint at Turn~0 and must issue follow-up questions across subsequent turns to gather enough evidence for a diagnosis. The dialogue terminates when the examiner produces a terminal diagnosis or a maximum turn limit is reached. Every turn is logged to a structured JSON record: prompts, raw text, examiner question, patient utterance selected, and \cuis extracted from both sides.  CUI extraction applied to examiner outputs proceeds using the same two-stage pipeline detailed in Section \ref{sec:anchor_pv}, and enables the metrics described in Section~\ref{sec:discussion} over ontological identifiers, rather than surface forms or string matching.

\renewcommand{\thesubsection}{\thesection.\arabic{subsection}}

\vspace{-5pt}
\section{Discussion}
\label{sec:discussion}
\vspace{-5pt}
The primary advantage of MTDiag is its size, variety of sources, and well-defined schema and strict ontological anchoring, which allows
editing, masking, noise injection, and generation of different "patient profiles" from a single case. With that, it allows exploration of different areas of the search space. 
\paragraph{Demographic Fairness}
Separating clinical facts from the \texttt{Demographic} block in the \pv allows selective masking or altering of variables (e.g., age, race, insurance) during utterance generation to detect systemic undertriage or biased diagnostic trajectories for otherwise identical clinical presentations.
\paragraph{Multilingual Assessment}
Because UMLS \cui identifiers are language-agnostic, \pv strings can be swapped into any of 31 supported languages. Though initially English-focused, the schema natively supports multilingual dataset generation and dialogue simulation (via language-specific UserLMs) while preserving clinical ground truth.
\paragraph{Long-Tail Diagnostics}
Stratifying by source tests LLM robustness across the frequency spectrum: from routine, high-frequency ED triage (MIMIC-IV) to rare, complex cases (AJCR). Subsetting by specific ground-truth diagnoses further enables targeted, disease-specific "LLM clinical aptitude" assessments.
\newline
In addition, because every symptom in the \pv is mapped to a UMLS \cui, and every diagnosis to an ICD-10 code, we move beyond surface-level string matching. This allows us to automatically compute clinically meaningful metrics from dialogue logs and systematically evaluate how LLMs reason, where they fail, and how they interact. 
Using the \cui logs extracted from the examiner's questions at each turn, we propose the following metrics (Figure \ref{fig:worked_example}).
\paragraph{Semantic Diagnostic Distance}
Beyond binary accuracy ($\mathrm{Diag_{acc}}$), ICD-10 anchoring allows us to measure categorical severity. \texttt{G44.2} (Chapter G, nervous system) and \texttt{M31.6} (Chapter M, musculoskeletal system) sit in entirely different clinical chapters. A model outputting \texttt{I21.0} (acute myocardial infarction) when the ground truth is \texttt{I21.02} (STEMI) commits a minor underspecification; Dialogue B commits a severe categorical miss.

\paragraph{Symptom Elicitation Score ($S_\mathrm{dise}$)}
This metric assesses whether the model elicited the clinical evidence required for the diagnosis. Let $S_d$ be the set of symptoms associated with disease $d$ (drawn from the Human Phenotype Ontology, HPO), and $S_\mathrm{diag}$ be the \cui-resolved set of symptoms the examiner inquired about. We calculate a frequency-weighted relevance score:
$$f_w(s, S_d) = \omega(s, d) \cdot \mathbf{1}[s \in S_d]$$
where $\omega(s, d) \in \{1.0, 0.75, 0.5, 0.25, 0.1\}$ corresponds to HPO frequency qualifiers (from \textit{obligate} to \textit{very rare}).
\paragraph{Reliability and Bias}
A model that reaches the correct diagnosis without eliciting enough evidence, or its \textit{sine qua non} (strictly necessary) or pathognomonic (strictly sufficient) symptoms, commits a \textbf{diagnostic hallucination}. \cui anchoring makes logical errors automatically computable:
\begin{itemize}[noitemsep,topsep=0pt]
    \item \textbf{Reliability Score ($R_\mathrm{score}$):} A model is credited only if it reached the correct diagnosis \textit{and} its symptom elicitation score ($S_\mathrm{dise}$) exceeds a minimum threshold. 
    \item \textbf{Anchoring Bias:} Queried symptoms are exclusively mapped to a single incorrect candidate diagnosis (e.g., Dialogue B asking only migraine-associated questions).
    \item \textbf{Premature Closure:} A model commits to a terminal diagnosis before crossing a minimum threshold of necessary symptom elicitation for the diagnosis.
\end{itemize}
\paragraph{The Harm Index}
Diagnostic errors carry varying risk. Harm is clinically categorized into three tiers:
\begin{itemize}[noitemsep,topsep=0pt]
    \item \textbf{Tier 1 (Direct Harm):} Recommending a treatment contraindicated for the ground-truth diagnosis.
    \item \textbf{Tier 2 (Delay of Care):} Diagnosing a low-acuity condition when the ground truth is a time-sensitive emergency (e.g., Dialogue B missing GCA, risking irreversible bilateral blindness).
    \item \textbf{Tier 3 (Instructional Failure):} Omitting critical advice that the case warrants.
\end{itemize}
Importantly, these metrics are not specific to MTDiag, rather the ontology-grounded metric suite is a dialogue-level evaluation protocol that any system producing CUI-annotated logs can be scored against. More details and additional metrics are discussed in Appendix \ref{app:metrics}.

MTDiag's grounding in clinical data with structured ontological anchoring and decoupled utterance generation means the evaluation substrate is a generative pipeline over structured cases, not a set of static files that can be leaked/memorized. With this work we aim to motivate the design of transparent, reproducible, and knowledge-grounded benchmarks for critical dialogue-based tasks, and continued research into interoperable knowledge bases like the UMLS.
Code and dataset available via \href{https://github.com/piachouaifaty/MTDiag}{\texttt{github.com/piachouaifaty/MTDiag}}.

\section*{Limitations}

\paragraph{The UMLS as a Living Resource}
Ontological anchoring is both a significant asset and a source of experimental overhead. Our development, experiments, and working pipeline are tagged to the 2025AB release of the UMLS. Naturally, rerunning the anchor resolution pipeline on a different UMLS installation may result in different results, or \cui assignments, as the UMLS is a regularly updated, living resource. Although this may introduce set-up overhead, it is also advantageous in that one may perform a bespoke UMLS installation, and include/exclude certain vocabularies for compatibility with specific health systems. 

\paragraph{Data Governance and Release Constraints}
MIMIC-IV, though formally de-identified, is considered potentially re-identifiable under PhysioNet's Data Use Agreement, which mandates Zero Data Retention (ZDR) and prohibits sharing with third parties. MIMIC-derived portions of MTDiag will be released exclusively via PhysioNet under the standard credentialed access agreement, and all processing and evaluation involving MIMIC data must be conducted in a ZDR-compliant environment. 

\paragraph{AJCR Pretraining Contamination}
Case reports from the AJCR series are widely indexed and freely available online, meaning some portion is likely present in the pretraining corpora of evaluated models. We treat this as a point of inquiry rather than a disqualifier: the structured \pv generated from each report produces dialogue substantially different from the original prose, and future work can directly test whether models perform better on this subset relative to others.

\paragraph{Utterance Pool Coverage}
The current release provides a set of canonical utterances per symptom per \pv entry. This is a conservative starting point: real patients vary enormously in how they describe the same symptom. The pipeline supports full variant expansion across persona parameters, masking policies, health literacy levels, and noise injections (described in Appendix~\ref{app:utterance_details}), but generating and validating this expansion at scale is left to future work.

\paragraph{Multi-Model Evaluation}
This work introduces the pipeline and evaluation framework, and provides a validated dataset. Systematic multi-model "as an examiner" evaluation is the natural next step and is left to future work.

\paragraph{Patient-Chatbot Realism Revisited} As described in Section \ref{sec:patientchatbotrealism}, we take steps to ensure the realism of our selected cases and design. However, the necessity of crafting system prompts properly instructing the LLM to act like a medical professional is far removed from the reality of a user prompting an LLM. In real life, a user will simply begin a conversation, rarely setting careful inference parameters, and may have additional, unrelated context, and, potentially, user-defined settings that alter outputs. A completely realistic assessment of LLMs as medical assistants to real users would require a large-scale ontological mapping of highly unstructured, real user chatlogs and the application of similar metrics as the ones we introduce.

\section*{Ethical Considerations}
It is important to point out that the hard technical limits of LLM context windows and their "as-assistant" sycophantism combined with the tendency of patients to be imperfect subjects (hedging, minimizing, re-framing) \cite{redelmeier2001problems, ijas2010patient} calls into question the suitability of LLMs for long-range diagnostic tasks.
Physicians undergo years of study, specialized training, mentorship, and practical clinical experience, and receive input from other similarly qualified doctors, nurses, and practitioners while considering a diagnosis. Moreover, they are human beings with a quasi-unlimited "context window," equipped with empathy, intuition, and a spirit of discernment. Most importantly, doctors take an oath to do no harm. 
We therefore present this work as a more rigorous LLM-as-diagnostician evaluation paradigm compared to existing approaches, and highly encourage the use of MTDiag for systematic multi-model evaluation in order to identify LLM failure modes and carefully consider their autonomous deployment in situations that may lead to the harming or loss of human life.

\section*{Acknowledgments}
The authors gratefully acknowledge the computing time granted by Federal Ministry of Education and Research (BMBF) project AI service center KISSKI (Grant N. 01IS22093A-E). The calculations for this research were conducted with computing resources under the project \textit{Towards a Realistic Multi-turn Medical Dialogue Benchmark for LLMs}. 
The authors also thank Dr. Maja Miličić Brandt for her ontology expertise, illuminating discussions around the UMLS, and meticulous feedback on the manuscript.


\bibliography{custom}

@article{gong2025knowledge,
  title={Knowledge-practice performance gap in clinical large language models: systematic review of 39 benchmarks},
  author={Gong, Eun Jeong and Bang, Chang Seok and Lee, Jae Jun and Baik, Gwang Ho},
  journal={Journal of Medical Internet Research},
  volume={27},
  pages={e84120},
  year={2025},
  publisher={JMIR Publications Toronto, Canada}
}

@article{vladika2026investigating,
  title={Investigating expectations and needs regarding the use of large language models at Bavarian university clinics},
  author={Vladika, Juraj and Fichtl, Alexander and Matthes, Florian},
  journal={Scientific Reports},
  year={2026},
  publisher={Nature Publishing Group UK London}
}

@Article{medspacy,
   Author="Eyre, H.  and Chapman, A. B.  and Peterson, K. S.  and Shi, J.  and Alba, P. R.  and Jones, M. M.  and Box, T. L.  and DuVall, S. L.  and Patterson, O. V. ",
   Title="{{L}aunching into clinical space with medspa{C}y: a new clinical text processing toolkit in {P}ython}",
   Journal="AMIA Annu Symp Proc",
   Year="2021",
   Volume="2021",
   Pages="438--447"
}

@inproceedings{jin2019pubmedqa,
  title={PubMedQA: A Dataset for Biomedical Research Question Answering},
  author={Jin, Qiao and Dhingra, Bhuwan and Liu, Zhengping and Cohen, William and Lu, Xinghua},
  booktitle={Proceedings of the 2019 Conference on Empirical Methods in Natural Language Processing and the 9th International Joint Conference on Natural Language Processing (EMNLP-IJCNLP)},
  pages={2567--2577},
  year={2019}
}

@inproceedings{nentidis2025bioasq,
  title={Overview of BioASQ 2025: The thirteenth BioASQ challenge on large-scale biomedical semantic indexing and question answering},
  author={Nentidis, Anastasios and Katsimpras, Georgios and Krithara, Anastasia and Krallinger, Martin and Rodr{\'\i}guez-Ortega, Miguel and Rodriguez-L{\'o}pez, Eduard and Loukachevitch, Natalia and Sakhovskiy, Andrey and Tutubalina, Elena and Dimitriadis, Dimitris and others},
  booktitle={International Conference of the Cross-Language Evaluation Forum for European Languages},
  pages={173--198},
  year={2025},
  organization={Springer}
}

@article{jin2021disease,
  title={What disease does this patient have? a large-scale open domain question answering dataset from medical exams},
  author={Jin, Di and Pan, Eileen and Oufattole, Nassim and Weng, Wei-Hung and Fang, Hanyi and Szolovits, Peter},
  journal={Applied Sciences},
  volume={11},
  number={14},
  pages={6421},
  year={2021},
  publisher={MDPI}
}

@inproceedings{pal2022medmcqa,
  title={Medmcqa: A large-scale multi-subject multi-choice dataset for medical domain question answering},
  author={Pal, Ankit and Umapathi, Logesh Kumar and Sankarasubbu, Malaikannan},
  booktitle={Conference on health, inference, and learning},
  pages={248--260},
  year={2022},
  organization={PMLR}
}

@inproceedings{alaa2025position,
  title={Position: Medical large language model benchmarks should prioritize construct validity},
  author={Alaa, Ahmed and Hartvigsen, Thomas and Golchini, Niloufar and Dutta, Shiladitya and Dean, Frances and Raji, Inioluwa Deborah and Zack, Travis},
  booktitle={Forty-second International Conference on Machine Learning Position Paper Track},
  year={2025}
}

@article{tu2025towards,
  title={Towards conversational diagnostic artificial intelligence},
  author={Tu, Tao and Schaekermann, Mike and Palepu, Anil and Saab, Khaled and Freyberg, Jan and Tanno, Ryutaro and Wang, Amy and Li, Brenna and Amin, Mohamed and Cheng, Yong and others},
  journal={Nature},
  volume={642},
  number={8067},
  pages={442--450},
  year={2025},
  publisher={Nature Publishing Group UK London}
}

@inproceedings{fan-etal-2025-ai,
    title = "{AI} Hospital: Benchmarking Large Language Models in a Multi-agent Medical Interaction Simulator",
    author = "Fan, Zhihao  and
      Wei, Lai  and
      Tang, Jialong  and
      Chen, Wei  and
      Siyuan, Wang  and
      Wei, Zhongyu  and
      Huang, Fei",
    editor = "Rambow, Owen  and
      Wanner, Leo  and
      Apidianaki, Marianna  and
      Al-Khalifa, Hend  and
      Eugenio, Barbara Di  and
      Schockaert, Steven",
    booktitle = "Proceedings of the 31st International Conference on Computational Linguistics",
    month = jan,
    year = "2025",
    address = "Abu Dhabi, UAE",
    publisher = "Association for Computational Linguistics",
    url = "https://aclanthology.org/2025.coling-main.680/",
    pages = "10183--10213"
}

@inproceedings{johri2024craft,
  title={CRAFT-MD: A conversational evaluation framework for comprehensive assessment of clinical LLMs},
  author={Johri, Shreya and Jeong, Jaehwan and Tran, Benjamin A and Schlessinger, Daniel I and Wongvibulsin, Shannon and Cai, Zhuo Ran and Daneshjou, Roxana and Rajpurkar, Pranav},
  booktitle={AAAI 2024 Spring Symposium on Clinical Foundation Models},
  year={2024}
}

@article{fansi2022ddxplus,
  title={DDXPlus: A new dataset for automatic medical diagnosis},
  author={Fansi Tchango, Arsene and Goel, Rishab and Wen, Zhi and Martel, Julien and Ghosn, Joumana},
  journal={Advances in neural information processing systems},
  volume={35},
  pages={31306--31318},
  year={2022}
}

@article{draelos2026large,
  title={Large language models provide unsafe answers to patient-posed medical questions},
  author={Draelos, Rachel L and Afreen, Samina and Blasko, Barbara and Brazile, Tiffany L and Chase, Natasha and Desai, Dimple Patel and Evert, Jessica and Gardner, Heather L and Herrmann, Lauren and House, Aswathy Vaikom and others},
  journal={NPJ digital medicine},
  volume={9},
  number={1},
  pages={241},
  year={2026},
  publisher={Nature Publishing Group UK London}
}

@article{PhysioNet-mimiciv-3.1,
  author = {Johnson, Alistair and Bulgarelli, Lucas and Pollard, Tom and Gow, Brian and Moody, Benjamin and Horng, Steven and Celi, Leo Anthony and Mark, Roger},
  title = {{MIMIC-IV}},
  journal = {{PhysioNet}},
  year = {2024},
  month = oct,
  note = {Version 3.1},
  doi = {10.13026/kpb9-mt58},
  url = {https://doi.org/10.13026/kpb9-mt58}
}

@article{johnson2023mimic,
  title={MIMIC-IV, a freely accessible electronic health record dataset},
  author={Johnson, Alistair EW and Bulgarelli, Lucas and Shen, Lu and Gayles, Alvin and Shammout, Ayad and Horng, Steven and Pollard, Tom J and Hao, Sicheng and Moody, Benjamin and Gow, Brian and others},
  journal={Scientific data},
  volume={10},
  number={1},
  pages={1},
  year={2023},
  publisher={Nature Publishing Group UK London}
}

@article{hirosawa2024comparative,
  title={Comparative study to evaluate the accuracy of differential diagnosis lists generated by gemini advanced, gemini, and bard for a case report series analysis: cross-sectional study},
  author={Hirosawa, Takanobu and Harada, Yukinori and Tokumasu, Kazuki and Ito, Takahiro and Suzuki, Tomoharu and Shimizu, Taro},
  journal={JMIR Medical Informatics},
  volume={12},
  pages={e63010},
  year={2024},
  publisher={JMIR Publications Toronto, Canada}
}

@article{bodenreider2004unified,
  title={The unified medical language system (UMLS): integrating biomedical terminology},
  author={Bodenreider, Olivier},
  journal={Nucleic acids research},
  volume={32},
  number={suppl\_1},
  pages={D267--D270},
  year={2004},
  publisher={Oxford University Press}
}

@article{naous2025flipping,
  title={Flipping the dialogue: Training and evaluating user language models},
  author={Naous, Tarek and Laban, Philippe and Xu, Wei and Neville, Jennifer},
  journal={arXiv preprint arXiv:2510.06552},
  year={2025}
}

@article{sellergren2025medgemma,
  title={Medgemma technical report},
  author={Sellergren, Andrew and Kazemzadeh, Sahar and Jaroensri, Tiam and Kiraly, Atilla and Traverse, Madeleine and Kohlberger, Timo and Xu, Shawn and Jamil, Fayaz and Hughes, C{\'\i}an and Lau, Charles and others},
  journal={arXiv preprint arXiv:2507.05201},
  year={2025}
}

@article{xu2024benchmarking,
  title={Benchmarking benchmark leakage in large language models},
  author={Xu, Ruijie and Wang, Zengzhi and Fan, Run-Ze and Liu, Pengfei},
  journal={arXiv preprint arXiv:2404.18824},
  year={2024}
}

@inproceedings{balloccu-etal-2024-leak,
    title = "Leak, Cheat, Repeat: Data Contamination and Evaluation Malpractices in Closed-Source {LLM}s",
    author = "Balloccu, Simone  and
      Schmidtov{\'a}, Patr{\'i}cia  and
      Lango, Mateusz  and
      Dusek, Ondrej",
    editor = "Graham, Yvette  and
      Purver, Matthew",
    booktitle = "Proceedings of the 18th Conference of the European Chapter of the Association for Computational Linguistics (Volume 1: Long Papers)",
    month = mar,
    year = "2024",
    address = "St. Julian{'}s, Malta",
    publisher = "Association for Computational Linguistics",
    url = "https://aclanthology.org/2024.eacl-long.5/",
    doi = "10.18653/v1/2024.eacl-long.5",
    pages = "67--93"
}

@article{redelmeier2001problems,
  title={Problems for clinical judgement: 1. Eliciting an insightful history of present illness},
  author={Redelmeier, Donald A and Schull, Michael J and Hux, Janet E and Tu, Jack V and Ferris, Lorraine E},
  journal={Cmaj},
  volume={164},
  number={5},
  pages={647--651},
  year={2001},
  publisher={CMAJ}
}

@article{sonnenberg2002translating,
  title={Translating vague complaints into precise symptoms: the implications of a poor medical history},
  author={Sonnenberg, Amnon and Gogel, Howard K},
  journal={European journal of gastroenterology \& hepatology},
  volume={14},
  number={3},
  pages={317--321},
  year={2002},
  publisher={LWW}
}

@article{ijas2010patient,
  title={Patient resistance towards diagnosis in primary care: Implications for concordance},
  author={Ij{\"a}s-Kallio, Taru and Ruusuvuori, Johanna and Per{\"a}kyl{\"a}, Anssi},
  journal={Health:},
  volume={14},
  number={5},
  pages={505--522},
  year={2010},
  publisher={SAGE Publications Sage UK: London, England}
}

@article{hager2024evaluation,
  title={Evaluation and mitigation of the limitations of large language models in clinical decision-making},
  author={Hager, Paul and Jungmann, Friederike and Holland, Robbie and Bhagat, Kunal and Hubrecht, Inga and Knauer, Manuel and Vielhauer, Jakob and Makowski, Marcus and Braren, Rickmer and Kaissis, Georgios and others},
  journal={Nature medicine},
  volume={30},
  number={9},
  pages={2613--2622},
  year={2024},
  publisher={Nature Publishing Group US New York}
}

@inproceedings{arias2025automatic,
  title={Automatic evaluation of healthcare LLMs beyond question-answering},
  author={Arias-Duart, Anna and Martin-Torres, Pablo Agustin and Hinjos, Daniel and Bernabeu-Perez, Pablo and Ganzabal, Lucia Urcelay and Mallo, Marta Gonzalez and Gururajan, Ashwin Kumar and Lopez-Cuena, Enrique and Alvarez-Napagao, Sergio and Garcia-Gasulla, Dario},
  booktitle={Proceedings of the 2025 Conference of the Nations of the Americas Chapter of the Association for Computational Linguistics: Human Language Technologies (Volume 2: Short Papers)},
  pages={108--130},
  year={2025}
}

@article{vally2023errors,
  title={Errors in clinical diagnosis: a narrative review},
  author={Vally, Zunaid Ismail and Khammissa, Razia AG and Feller, Gal and Ballyram, Raoul and Beetge, Michaela and Feller, Liviu},
  journal={Journal of International Medical Research},
  volume={51},
  number={8},
  pages={03000605231162798},
  year={2023},
  publisher={SAGE Publications Sage UK: London, England}
}

@article{heston2024chatgpt,
  title={ChatGPT provides inconsistent risk-stratification of patients with atraumatic chest pain},
  author={Heston, Thomas F and Lewis, Lawrence M},
  journal={PLoS One},
  volume={19},
  number={4},
  pages={e0301854},
  year={2024},
  publisher={Public Library of Science San Francisco, CA USA}
}

@article{ramaswamy2026chatgpt,
  title={ChatGPT Health performance in a structured test of triage recommendations},
  author={Ramaswamy, Ashwin and Tyagi, Alvira and Hugo, Hannah and Jiang, Joy and Jayaraman, Pushkala and Jangda, Mateen and Te, Alexis E and Kaplan, Steven A and Lampert, Joshua and Freeman, Robert and others},
  journal={Nature Medicine},
  pages={1--5},
  year={2026},
  publisher={Nature Publishing Group}
}

@article{kearney2025language,
  title={Language models change facts based on the way you talk},
  author={Kearney, Matthew and Binns, Reuben and Gal, Yarin},
  journal={arXiv preprint arXiv:2507.14238},
  year={2025}
}

@inproceedings{zhou2025unveiling,
  title={Unveiling performance challenges of large language models in low-resource healthcare: A demographic fairness perspective},
  author={Zhou, Yue and Di Eugenio, Barbara and Cheng, Lu},
  booktitle={Proceedings of the 31st International Conference on Computational Linguistics},
  pages={7266--7278},
  year={2025}
}

@article{macherla2023mddial,
  title={Mddial: A multi-turn differential diagnosis dialogue dataset with reliability evaluation},
  author={Macherla, Srija and Luo, Man and Parmar, Mihir and Baral, Chitta},
  journal={arXiv preprint arXiv:2308.08147},
  year={2023}
}

@article{sangwon2025evaluating,
  title={Evaluating Large Language Model Diagnostic Performance on JAMA Clinical Challenges via a Multi-Agent Conversational Framework},
  author={Sangwon, Karl L and Zhang, Jeff and Steele, Robert and Stryker, Jaden and Lee, Jin Vivian and Choi, Joanne and Vishwanath, Krithik and Alber, Daniel Alexander and Kondziolka, Douglas and Mankowski, Michal and others},
  journal={medRxiv},
  pages={2025--08},
  year={2025},
  publisher={Cold Spring Harbor Laboratory Press}
}

@inproceedings{zeng2020meddialog,
  title     = {{MedDialog}: Large-scale Medical Dialogue Datasets},
  author    = {Zeng, Guangtao and Yang, Wenmian and Ju, Zeqian and Yang, Yue and
               Wang, Sicheng and Zhang, Ruisi and Zhou, Meng and Zeng, Jiaqi and
               Dong, Xiangyu and Zhang, Ruoyu and Fang, Hongchao and Zhu, Penghui and
               Chen, Shu and Xie, Pengtao},
  booktitle = {Proceedings of the 2020 Conference on Empirical Methods in Natural
               Language Processing (EMNLP)},
  pages     = {9241--9250},
  year      = {2020},
  address   = {Online},
  publisher = {Association for Computational Linguistics},
  doi       = {10.18653/v1/2020.emnlp-main.743},
  url       = {https://aclanthology.org/2020.emnlp-main.743/}
}

@article{yan2021remedi,
  title   = {{ReMeDi}: Resources for Multi-domain, Multi-service, Medical Dialogues},
  author  = {Yan, Guojun and Pei, Jiahuan and Ren, Pengjie and Ren, Zhaochun and
             Xin, Xin and Liang, Huasheng and de Rijke, Maarten and Chen, Zhumin},
  journal = {arXiv preprint arXiv:2109.00430},
  year    = {2021},
  url     = {https://arxiv.org/abs/2109.00430},
  doi     = {10.48550/arXiv.2109.00430}
}

@article{liu2022meddg,
  title   = {{MedDG}: An Entity-Centric Medical Consultation Dataset for
             Entity-Aware Medical Dialogue Generation},
  author  = {Liu, Wenge and Tang, Jianheng and Cheng, Yi and Li, Wenjie and
             Zheng, Yefeng and Liang, Xiaodan},
  journal = {arXiv preprint arXiv:2010.07497},
  year    = {2022},
  url     = {https://arxiv.org/abs/2010.07497},
  doi     = {10.48550/arXiv.2010.07497}
}

@inproceedings{wei2018task,
  title     = {Task-oriented Dialogue System for Automatic Diagnosis},
  author    = {Wei, Zhongyu and Liu, Qianlong and Peng, Baolin and Tou, Huaixiao and
               Chen, Ting and Huang, Xuanjing and Wong, Kam-fai and Dai, Xiangying},
  booktitle = {Proceedings of the 56th Annual Meeting of the Association for
               Computational Linguistics (Volume 2: Short Papers)},
  pages     = {201--207},
  year      = {2018},
  address   = {Melbourne, Australia},
  publisher = {Association for Computational Linguistics},
  doi       = {10.18653/v1/P18-2033},
  url       = {https://aclanthology.org/P18-2033/}
}

@inproceedings{xu2019end,
  title     = {End-to-End Trainable Non-Collaborative Dialog System for Automatic
               Diagnosis},
  author    = {Xu, Lin and Zhou, Qixian and Gong, Ke and Liang, Xiaodan and
               Tang, Jianheng and Lin, Liang},
  booktitle = {Proceedings of the AAAI Conference on Artificial Intelligence},
  volume    = {33},
  number    = {01},
  pages     = {7346--7353},
  year      = {2019},
  doi       = {10.1609/aaai.v33i01.33017346},
  url       = {https://ojs.aaai.org/index.php/AAAI/article/view/4722}
}

@inproceedings{lin2019enhancing,
  title     = {Enhancing Dialogue Symptom Diagnosis with Global Attention and
               Symptom Graph},
  author    = {Lin, Xinzhu and He, Xiahui and Chen, Qin and Tou, Huaixiao and
               Wei, Zhongyu and Chen, Ting},
  booktitle = {Proceedings of the 2019 Conference on Empirical Methods in Natural
               Language Processing and the 9th International Joint Conference on
               Natural Language Processing (EMNLP-IJCNLP)},
  pages     = {5033--5042},
  year      = {2019},
  address   = {Hong Kong, China},
  publisher = {Association for Computational Linguistics},
  doi       = {10.18653/v1/D19-1508},
  url       = {https://aclanthology.org/D19-1508/}
}

@inproceedings{zhang2020mie,
  title     = {{MIE}: A Medical Information Extractor towards Medical Dialogues},
  author    = {Zhang, Yuanzhe and Jiang, Zhongtao and Zhang, Tao and Liu, Shiwan and
               Cao, Jiarun and Liu, Kang and Liu, Shengping and Zhao, Jun},
  booktitle = {Proceedings of the 58th Annual Meeting of the Association for
               Computational Linguistics},
  pages     = {6460--6469},
  year      = {2020},
  address   = {Online},
  publisher = {Association for Computational Linguistics},
  doi       = {10.18653/v1/2020.acl-main.576},
  url       = {https://aclanthology.org/2020.acl-main.576/}
}

@article{szydelko2022arteritic,
  title={Arteritic anterior ischemic optic neuropathy in the course of giant cell arteritis after COVID-19},
  author={Szyde{\l}ko-Pa{\'s}ko, Urszula and Prze{\'z}dziecka-Do{\l}yk, Joanna and Kr{\k{e}}cicka, Julia and Ma{\l}ecki, Rafa{\l} and Misiuk-Hoj{\l}o, Marta and Turno-Kr{\k{e}}cicka, Anna},
  journal={The American Journal of Case Reports},
  volume={23},
  pages={e933471--1},
  year={2022}
}

@article{soldaini2016quickumls,
  title={QuickUMLS: a fast, unsupervised approach for medical concept extraction},
  author={Soldaini, Luca and Goharian, Nazli},
  year={2016},
  url={https://ir.cs.georgetown.edu/downloads/quickumls.pdf}
}

@article{Simmons2025Extracting,
  author  = {Simmons, A and Takkavatakarn, K and McDougal, M and Dilcher, B and Pincavitch, J and Meadows, L and Kauffman, J and Klang, E and Wig, R and Smith, G and Soroush, A and Freeman, R and Apakama, D. J. and Charney, A. W. and Kohli-Seth, R and Nadkarni, G. N. and Sakhuja, A},
  title   = {Extracting International Classification of Diseases Codes from Clinical Documentation Using Large Language Models},
  journal = {Applied Clinical Informatics},
  year    = {2025},
  volume  = {16},
  number  = {2},
  pages   = {337--344},
  doi     = {10.1055/a-2491-3872},
  pmid    = {39608761},
  pmcid   = {PMC12020521}
}

@article{klang2024assessing,
  title={Assessing retrieval-augmented large language model performance in emergency department ICD-10-CM coding compared to human coders},
  author={Klang, Eyal and Tessler, Idit and Apakama, Donald U and Abbott, Ethan and Glicksberg, Benjamin S and Arnold, Monique and Moses, Akini and Sakhuja, Ankit and Soroush, Ali and Charney, Alexander W and others},
  journal={medRxiv},
  year={2024}
}

\appendix

\section{Datasets Considered but Excluded}
\label{app:unuseddatasets}
\vspace{-5pt}
Several alternative dialogue corpora were evaluated and excluded for the following reasons:

\begin{itemize}[noitemsep,topsep=0pt]
    \item \textbf{Lack of multi-turn diagnostic structure:} MedDialog \citep{zeng2020meddialog} consists primarily of single-turn, forum-style QA lacking a diagnostic trajectory. MIE \citep{zhang2020mie} is an information-extraction corpus, not a dialogue dataset.
    \item \textbf{Rigid generation:} MDDial \citep{macherla2023mddial} relies on template-driven generation, lacking naturalistic variation and sufficient multi-turn depth.
    \item \textbf{Language constraints:} ReMeDi \citep{yan2021remedi}, MedDG \citep{liu2022meddg}, and MDD \citep{wei2018task} are exclusively Chinese-language corpora.
    \item \textbf{Narrow clinical scope:} DX \citep{xu2019end} and the dataset by \citet{lin2019enhancing} are heavily restricted in scale, covering only five pediatric conditions.
\end{itemize}
\vspace{-5pt}

\section{Source Datasets}
\label{app:sourcedatasets}
 \vspace{-5pt}

\subsection{DDXPlus}
\label{app:ddxplus}

\subsubsection{Evidence Structure}
\vspace{-5pt}

Evidences are typed into three categories. 
\paragraph{Binary} (\texttt{B}) evidences represent 
yes/no symptom presence. 
\paragraph{Categorical} (\texttt{C}) evidences encode a single value 
from a fixed set of options (e.g. pain character, onset pattern). 
\paragraph{Multi-choice} 
(\texttt{M}) evidences are child items parented to a categorical evidence via a 
\texttt{code\_question} hierarchy, allowing a patient to affirm multiple sub-features of a 
single symptom simultaneously (e.g. multiple pain locations). 

Evidences are further annotated 
as \texttt{is\_antecedent} to distinguish presenting symptoms from medical history items.

\subsubsection{Preprocessing and Reinterpretation}

The \texttt{default\_value} of 0 on binary evidences does not encode an explicit patient denial; 
it reflects the absence of synthesis during dataset generation. Under the closed-world assumption, 
presence in \texttt{EVIDENCES} means affirmed (value = 1); absence means negative (value = 0). 
Evidence states are therefore reconstructed directly from list membership rather than from 
default values, preventing misrepresentation of negative findings.

Each patient in DDXPlus carries an \texttt{INITIAL\_EVIDENCE} field, a randomly selected 
binary evidence used solely to bootstrap the original rule-based ADD system, with no 
inherent clinical salience. We retain it as the opening chief complaint, both to preserve 
the original diagnostic progression structure of the dataset and to ensure consistency 
across comparisons. The value of DDXPlus for this benchmark lies primarily in its  differential diagnosis trajectories and structured evidence instantiation rather than the realism of its opening utterance.
\vspace{-5pt}

\subsection{MIMIC-IV}
\label{app:mimic}
\label{app:mimic_modules}
\vspace{-5pt}

MIMIC-IV~\cite{PhysioNet-mimiciv-3.1} is a large, deidentified electronic health record database
sourced from the Beth Israel Deaconess Medical Center (BIDMC) in Boston, MA, covering patients
admitted to the emergency department or an intensive care unit between 2008 and 2019.
It contains data for over 65,000 ICU (Intensive Care Unit) stays and over 200,000 ED (Emergency Department) visits,
and adopts a modular architecture that facilitates both individual and combined use of
disparate data sources.  All patient identifiers are replaced with randomized
surrogates and dates are shifted by a patient-specific offset, preserving intra-patient
temporal consistency while preventing re-identification.
\vspace{-5pt}

\subsubsection{Modules}
Unlike traditional uses of MIMIC, involving extraction of dense longitudinal time-series from MIMIC-IV for predictive
modeling, our benchmark requires coherent narrative patient presentations. 
Our dataset makes use of four official MIMIC modules (hosp, ed) 
and two community extensions (bhc, cdn) (see Table \ref{tab:mimic_files}).
\vspace{-5pt}

\paragraph{MIMIC-IV v3.1 (Hosp)}~\cite{johnson2023mimic} is the core hospital module, covering
inpatient admissions at BIDMC. We extract five tables: \texttt{patients} (demographics
and date of death), \texttt{admissions} (admission type, insurance, discharge disposition,
and in-hospital mortality flag), \texttt{diagnoses\_icd} (ICD-coded discharge diagnoses
with sequence ordering reflecting clinical priority), \texttt{d\_icd\_diagnoses} (the ICD
code dictionary, used for crosswalk and label resolution), and \texttt{omr} (outpatient
measurements including height, weight, and BMI recorded across visits).
\vspace{-5pt}

\paragraph{MIMIC-IV-ED v2.2} extends MIMIC-IV to the emergency
department, covering over 200,000 ED visits. It is the entry point for our entire
extraction pipeline. We use \texttt{triage} (chief complaint, triage vital signs, pain
score, and ESI acuity level), \texttt{edstays} (ED stay timeline and crucially the
\texttt{hadm\_id} bridge key that links ED visits to hospital admissions, partitioning
our two-track cohort), \texttt{diagnosis} (ICD-coded ED-specific diagnoses, assigned
independently of the final hospital billing diagnoses and therefore serving as the Track
1 ground truth), \texttt{medrecon} (home medications recorded at triage), and
\texttt{vitalsign} (time-series vital sign measurements taken during the ED stay).
\vspace{-5pt}

\paragraph{MIMIC-IV-Note v2.2} provides 331,794 deidentified
free-text discharge summaries from 145,915 patients, with protected health information
replaced by three consecutive underscores following a hybrid rule-based and neural
deidentification approach. 
\vspace{-5pt}

\paragraph{BHC extension} (Labelled Notes: Hospital Course v1.2.0) is a
community-contributed PhysioNet dataset providing 125,123 parsed and labeled hospital
course narratives derived from \textbf{MIMIC-IV-Note v2.2} discharge summaries, keyed on \texttt{hadm\_id}. We use the presence and count of parsed hospital course
segments as a data richness signal to inform the patient-level deduplication priority
scheme for admitted cases, described later.
\vspace{-5pt}

\paragraph{MIMIC-IV-Ext CDM v1.1} is a curated community extension
covering 2,400 admitted cases across four abdominal pathologies: appendicitis (957),
cholecystitis (648), pancreatitis (538), and diverticulitis (257). Each case provides
pre-extracted, labeled, clinical data including HPI text, physical examination
findings, 138,788 laboratory results from 480 unique tests, 4,403 microbiology results,
and 5,959 radiology reports across CT, X-ray, ultrasound, and MRI modalities. Crucially,
the CDM authors explicitly exclude cases where the pathology name appears in the HPI
(indicating pre-diagnosed transfers) and strip diagnostic conclusions from radiology
findings sections, making the narratives genuinely suitable for diagnostic simulation.
Where a CDM record exists for an admitted case in our pipeline, it takes precedence over
the note-derived HPI as the Track 2 narrative input, given its dedicated curation and higher fidelity.
\vspace{-5pt}

\begin{table*}[htbp]
\centering
\resizebox{\textwidth}{!}{%
\begin{tabular}{@{}llll@{}}
\toprule
\textbf{Directory} & \textbf{Source} & \textbf{Docs} & \textbf{Key Columns} \\ \midrule

\multicolumn{4}{l}{\textbf{ed/} \textit{MIMIC-IV-ED v2.2}} \\
\quad \texttt{triage.csv.gz} & & \href{https://mimic.mit.edu/docs/iv/modules/ed/triage/}{Docs} & \texttt{subject\_id, stay\_id, chiefcomplaint, temperature, heartrate, resprate, o2sat, sbp, dbp, pain, acuity} \\
\quad \texttt{edstays.csv.gz} & & \href{https://mimic.mit.edu/docs/iv/modules/ed/edstays/}{Docs} & \texttt{subject\_id, hadm\_id, stay\_id, intime, outtime, arrival\_transport, disposition} \\
\quad \texttt{diagnosis.csv.gz} & & \href{https://mimic.mit.edu/docs/iv/modules/ed/diagnosis/}{Docs} & \texttt{stay\_id, seq\_num, icd\_code, icd\_version} \\
\quad \texttt{medrecon.csv.gz} & & \href{https://mimic.mit.edu/docs/iv/modules/ed/medrecon/}{Docs} & \texttt{stay\_id, name, gsn, ndc, etc\_rn} \\
\quad \texttt{vitalsign.csv.gz} & & \href{https://mimic.mit.edu/docs/iv/modules/ed/vitalsign/}{Docs} & \texttt{stay\_id, charttime, temperature, heartrate, resprate, o2sat, sbp, dbp, pain} \\
\addlinespace

\multicolumn{4}{l}{\textbf{hosp/} \textit{MIMIC-IV v3.1}} \\
\quad \texttt{patients.csv.gz} & & \href{https://mimic.mit.edu/docs/iv/modules/hosp/patients/}{Docs} & \texttt{subject\_id, gender, anchor\_age, dod} \\
\quad \texttt{admissions.csv.gz} & & \href{https://mimic.mit.edu/docs/iv/modules/hosp/admissions/}{Docs} & \texttt{subject\_id, hadm\_id, race, admittime, dischtime, admission\_type, discharge\_location, hospital\_expire\_flag} \\
\quad \texttt{diagnoses\_icd.csv.gz} & & \href{https://mimic.mit.edu/docs/iv/modules/hosp/diagnoses_icd/}{Docs} & \texttt{hadm\_id, seq\_num, icd\_code, icd\_version} \\
\quad \texttt{d\_icd\_diagnoses.csv.gz} & & \href{https://mimic.mit.edu/docs/iv/modules/hosp/d_icd_diagnoses/}{Docs} & \texttt{icd\_code, icd\_version, long\_title} \\
\quad \texttt{omr.csv.gz} & & \href{https://mimic.mit.edu/docs/iv/modules/hosp/omr/}{Docs} & \texttt{subject\_id, chartdate, seq\_num, result\_name, result\_value} \\
\addlinespace

\multicolumn{4}{l}{\textbf{note/} \textit{MIMIC-IV-Note v2.2}} \\
\quad \texttt{discharge.csv.gz} & & \href{https://mimic.mit.edu/docs/iv/modules/note/discharge/}{Docs} & \texttt{subject\_id, hadm\_id, note\_id, text} \\
\addlinespace

\multicolumn{4}{l}{\textbf{bhc/} \textit{Labelled Notes: Hospital Course v1.2.0}} \\
\quad \texttt{mimic-iv-bhc.csv} & \href{https://physionet.org/content/labelled-notes-hospital-course/1.2.0/}{PhysioNet} & --- & \texttt{hadm\_id, text} (parsed hospital course) \\
\addlinespace

\multicolumn{4}{l}{\textbf{cdn/} \textit{MIMIC-IV-Ext CDM v1.1} --- 2,400 abdominal pathology cases} \\
\quad \texttt{history\_of\_present\_illness.csv} & & --- & \texttt{hadm\_id, hpi} \\
\quad \texttt{physical\_examination.csv} & & --- & \texttt{hadm\_id, pe} \\
\quad \texttt{discharge\_diagnosis.csv} & & --- & \texttt{hadm\_id, discharge\_diagnosis} \\
\quad \texttt{laboratory\_tests.csv} & & --- & \texttt{hadm\_id, itemid, valuestr, ref\_range\_lower, ref\_range\_upper} \\
\quad \texttt{radiology\_reports.csv} & & --- & \texttt{hadm\_id, note\_id, modality, region, exam\_name, text} \\
\quad \texttt{microbiology.csv} & & --- & \texttt{hadm\_id, test\_itemid, valuestr, spec\_itemid} \\
\quad \texttt{icd\_diagnosis.csv} & & --- & \texttt{hadm\_id, icd\_diagnosis} \\
\quad \texttt{icd\_procedures.csv} & & --- & \texttt{hadm\_id, icd\_code, icd\_title, icd\_version} \\
\quad \texttt{discharge\_procedures.csv} & & --- & \texttt{hadm\_id, discharge\_procedure} \\
\bottomrule
\end{tabular}%
}
\caption{MIMIC-IV data file organization, sources, and extracted columns. }
\label{tab:mimic_files}
\end{table*}

\subsubsection{Population Selection and Clinical Framing}
\label{app:mimic_filtering}
\vspace{-5pt}

\begin{figure}
    \centering
    \includegraphics[width=\linewidth]{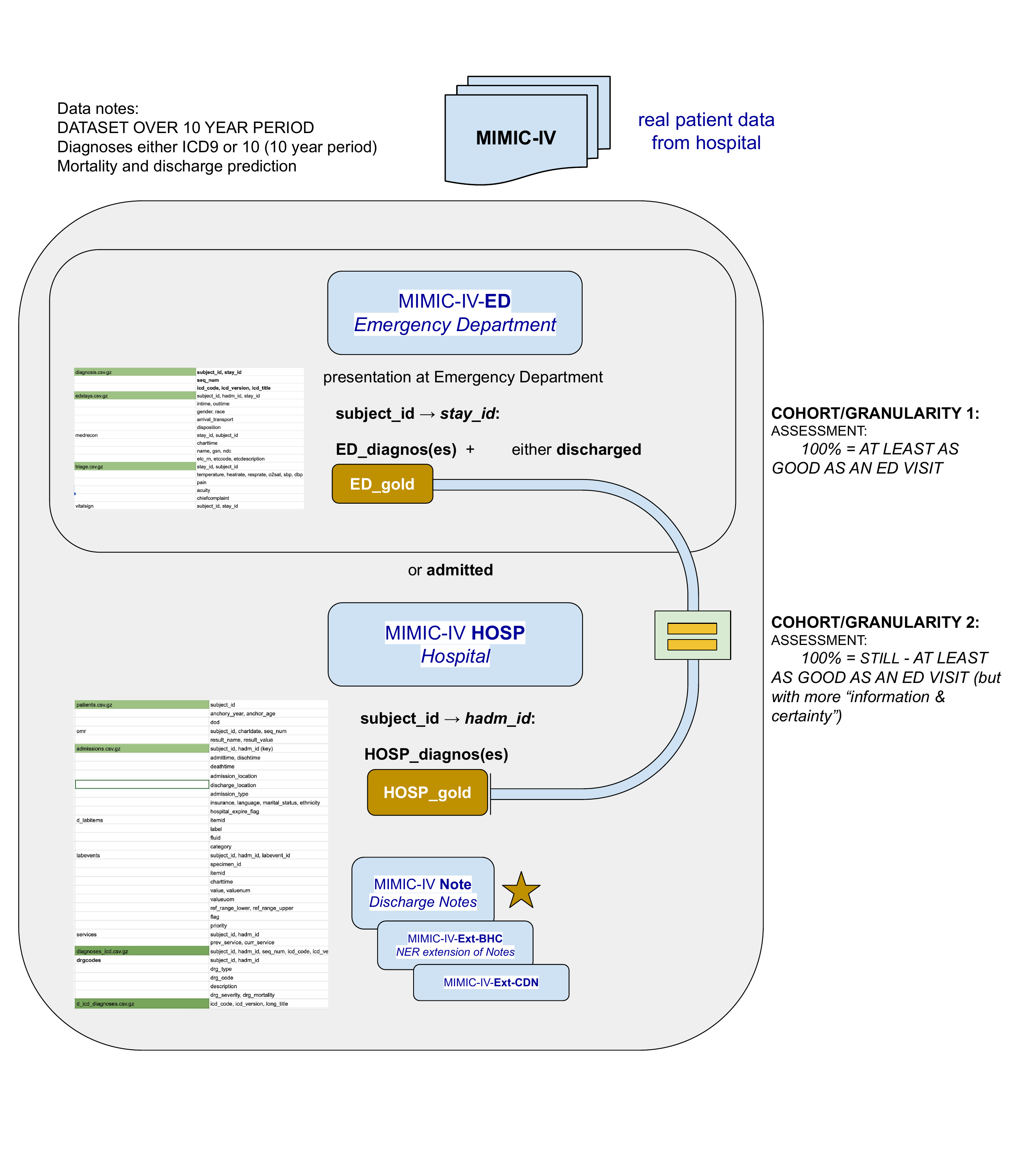}
    \caption{MIMIC-IV: Two Tracks/Cohorts}
    \label{fig:mimicfilering}
\end{figure}

The target population is patients who seek care by walking into an
emergency department of their own volition, and could plausibly
interact with an LLM instead. This framing
directly drives the cohort selection criteria. The filtering and selection pipeline is shown in Fig. \ref{fig:mimicfilering}.

The raw MIMIC-IV-ED dataset contains 425,087 ED visits. The first filtering step
excludes ambulance arrivals. In the United States, calling an ambulance is expensive and often reserved for acute emergencies; a patient arriving by ambulance is
by definition not self-presenting and is unlikely to consult a
chatbot as an alternative. This single filter removes a large proportion of the highest-acuity cases by proxy. Remaining high-acuity cases are addressed directly: visits with a triage pain
score above 8 out of 10 and visits assigned ESI acuity level 1 (the
``immediate/resuscitation'' tier of the Emergency Severity Index, a standardized 1--5
severity scale where 1 is most critical) are also excluded. Together these filters
operationalize the assumption that the benchmark should reflect cases where a patient
is coherent, able to communicate, and could realistically engage in a multi-turn dialogue with a chatbot.

A further quality filter removes cases with no diagnoses codes (ICD-10, or International Classification of Diseases 10th
Revision, is the international standard coding system for diagnoses), as ground truth
cannot be established for such cases. Finally, pain score entries recorded as
unresolvable free-text strings (``unable'', ``UTA'', ``critical'', and similar) are manually examined and dropped, indicating the patient was unable to communicate with the triage nurse about their pain, and thus falls outside our target population.

To ensure the integrity of the clinical data without reinventing complex data scrubbing rules, we selectively repurposed utility modules from an existing extensive MIMIC-IV processing pipeline. Specifically, we integrated their \texttt{outlier\_removal.py} functions to enforce physiological bounds on triage vitals (e.g., dropping artifacts/ sensor errors), their \texttt{uom\_conversion.py} scripts for unit standardization, and their ICD-9 to ICD-10 crosswalk mapping (MIMIC-IV data is recorded across 2 editions of ICD: 9 and 10).

\vspace{-5pt}

\subsubsection{Two-Tracks}
\vspace{-5pt}

After filtering, the remaining cohort naturally partitions into two clinically distinct
populations based on whether the ED visit resulted in hospital admission. Of the
425,087 raw visits, 222,071 (52.2\%) resulted in discharge without admission and
203,016 (47.8\%) in hospital admission. These two groups present fundamentally
different diagnostic challenges and map to different real-world scenarios:

\paragraph{Track 1: ED Discharge (telehealth use case):} The patient is
    evaluated and sent home. Feature space is restricted to ED data only (chief
    complaint, vital signs, ED questionnaire). Ground truth is the preliminary ED diagnosis assigned by
    the emergency physician at discharge.

\paragraph{Track 2: Hospital Admission (specialist/advanced diagnostic use
    case):} The patient is admitted for further workup. Triage data is augmented with
    a narrative History of Present Illness (HPI), the structured account of
    symptoms, onset, duration, and context recorded by the admitting physician, and, where available, structured clinical modalities from the BHC and CDM extensions.  Ground truth is the final verified discharge
    diagnosis, established after the full inpatient workup. 
    Note that for this track, we only retain cases where one or more of the hospital discharge diagnoses (after the patient is admitted and leaves) matches one or more of the preliminary ED diagnoses (the initial diagnos(e)s made by the ED physician which led to hospitalization).
    Since we use the patient's ED presentation for "chief complaint" and "initial patient state" as simulated "point of contact" with the examiner LLM, it would be unfair to expect the LLM to reach a diagnosis informed by a professional work-up after hospital admission, if that diagnosis was not sufficiently clear to the ED physicians upon presentation. Essentially, giving the chatbot a fair chance at the risk of introducing bias and skewing the data in its favor.
    
Track 1 tests whether a model can triage
from sparse initial information, while Track 2 tests deeper diagnostic reasoning
against a richer clinical picture. It also defines "confidence" in ground truth.

\vspace{-5pt}

\subsubsection{Data Extraction and Linkage}

Raw MIMIC-IV tables are preprocessed into a unified
424,995-row candidate dataframe covering 40 (sparse) features per visit. The pipeline
initialises from the ED module, merging \texttt{triage} and \texttt{edstays} via
\texttt{stay\_id} and \texttt{subject\_id} to form the base cohort. A
\texttt{LEFT JOIN} (retaining all rows from
(ED) table regardless of whether a matching record exists in the right (hospital)
table) is then executed to the hospital \texttt{admissions} and \texttt{patients}
tables on \texttt{hadm\_id} (Hospital Admission ID, the unique key linking an ED visit
to an inpatient admission). This join spine preserves non-admitted patients, who
receive a null \texttt{hadm\_id} and route to Track 1, while fully enriching admitted
patients (Track 2) with inpatient data. Subsequent steps pack time-series vital signs
(\texttt{vitalsign}), medication history (\texttt{medrecon}), and outpatient measurements
(\texttt{omr}) into per-visit aggregates; map both ED and hospital diagnoses to ICD-10
with ICD-9 crosswalk applied where needed; and join parsed
hospital course narratives from the BHC extension (125,123 records). Triage vital sign
outliers and unit inconsistencies are resolved using deterministic cleaning functions
from an existing MIMIC-IV pipeline.

For Track 2 cases where a CDM record exists, pre-extracted HPI text, physical
examination findings, laboratory results, radiology reports, and microbiology data are
merged in via \texttt{hadm\_id}, taking precedence over note-derived HPI given the
CDM's dedicated curation and pre-applied leakage controls. All downstream clinical
sections (``Hospital Course'', ``Assessment and Plan'', and similar) are masked
throughout the pipeline to prevent ground truth from entering the dialogue generation
stage.
\vspace{-5pt}

\subsubsection{Cohort Filtering Steps}
\vspace{-5pt}

The sequential filtering pipeline, applied after extraction, is summarised below.
Numbers reflect the state of each cohort at entry to each step.
\vspace{-5pt}

\paragraph{Pain score normalisation.}
Free-text pain entries that cannot be resolved to a numeric value are dropped, retained, or edited via a custom semi-manual process.
ED-only: 222,006~$\to$~218,902 ($-$3,104). Admitted: 60,847~$\to$~56,189
($-$4,658).
\vspace{-5pt}

\paragraph{Ambulance arrivals and high pain.}
Ambulance arrivals and pain scores above 8/10 are excluded.
ED-only: 218,902~$\to$~137,896 ($-$81,006). Admitted: 56,189~$\to$~23,245
($-$32,944).
\vspace{-5pt}

\paragraph{Triage acuity level 1.}
Immediate/resuscitation cases (ESI level 1) are excluded.
ED-only: 137,896~$\to$~135,623 ($-$2,273). Admitted: 23,245~$\to$~22,059
($-$1,186).
\vspace{-5pt}

\paragraph{Missing ICD-10 codes.}
Cases missing both primary and all secondary ICD-10 diagnosis codes are dropped, as
ground truth cannot be established.
ED-only: 135,623~$\to$~74,761 ($-$60,862). Admitted: 22,059~$\to$~22,059 ($-$0).
\vspace{-5pt}

\subsubsection{Patient-Level Deduplication}

MIMIC-IV contains multiple visits per patient. To prevent "frequent flyers" (patients who return to the hospital recurrently) from dominating the benchmark, and to ensure an i.i.d benchmark, a hierarchical deduplication strategy is applied. For patients with multiple visits,
each visit is scored by data richness:

\begin{itemize}
    \item \textbf{Priority 3} (1,460 visits): CDM record present \textit{and} parsed
    hospital course available - the richest possible entry
    \item \textbf{Priority 2} (185 visits): CDM record present, no hospital course
    \item \textbf{Priority 1} (8,473 visits): hospital course available, no CDM record
    \item \textbf{Priority 0} (8,940 visits): fallback - earliest chronological visit
\end{itemize}

The highest-priority visit per patient is retained.

\subsection{AJCR Case Reports}
\label{app:ajcr_schema}
\vspace{-5pt}
The original (\cite{hirosawa2024comparative}) dataset is a spreadsheet of 392 rows, each containing a PMID, a
DOI, up to three gold-standard diagnosis fields (\texttt{Final diagnosis1},
\texttt{Final diagnosis2}, \texttt{final diagnosis34}), and ranked 10-item differential
diagnosis lists generated by three LLMs at the time of the original study (Gemini
Advanced, Gemini, and Bard), (unused by us). We transform this into a rich structured dataset through a
two-stage pipeline. In the first stage, a custom scraping module normalizes each DOI
(e.g.\ \texttt{10.12659/AJCR.937787}) into a consistent identifier
(\texttt{AJCR937787}), constructs the corresponding URL via \texttt{doi.org}, and
retrieves the full-text HTML of the article, saving it to a per-case directory. Of the
392 cases, 391 were retrieved successfully; 1 failed due to an unavailable page and is
excluded. In the second stage, a specialised parser built on \texttt{BeautifulSoup}
maps the HTML DOM into a standardized nested JSON schema. The parser extracts:
granular metadata (title, authors, DOI, journal citation, publication date, and article
type) from HTML meta tags and structured page elements; the abstract segmented into
\textit{Background}, \textit{Case Report}, and \textit{Conclusions} fields via regex,
with a fallback to the page's \texttt{meta description} tag where structured paragraph
IDs are absent; keywords from the \texttt{meta keywords} tag; the full body text split
by section (Background, Discussion, Conclusions); figures catalogued by their DOM image
anchors with label and caption; and tables detected via their DOM anchor IDs.

Tables in AJCR articles exist in two formats. Where the table is rendered as a native
HTML \texttt{<table>} element, it is parsed using \texttt{pandas} and stored as both a
structured record list and a Markdown string representation. Where the table is rendered
as an image (a common occurrence in AJCR, where authors submit tables as figures),
it is stored with its source URL and caption. In both cases the table is stored
uniformly as a string representation within the JSON schema, ensuring
downstream access by all LLMs (and not just multi-modal ones for the to tables-as-images).

A key parsing challenge is the disambiguation of case series, in which a single
publication describes two or more distinct patients under one DOI. Simple title-string
matching for ``case series'' is insufficiently robust; instead, the parser detects case
series by identifying numbered section headers (e.g.\ ``Case 1'', ``Case 2'',
``CASE REPORT 1:'') within the case report body using a regex pattern, splitting the
narrative at these boundaries and assigning paragraphs, figures, and tables to their
respective patient. Figures and tables are resolved through named mention in the original HTML markdown. Figures and tables can be referenced by multiple cases and are
duplicated accordingly. Assets not referenced by any individual case (e.g.\ figures
appearing only in the Discussion section) are collected in a \texttt{general} block.
Each individual patient is emitted as a distinct case entry with its own
\texttt{case\_id} (e.g.\ \texttt{AJCR937787\_1}, \texttt{AJCR937787\_2} for a case
series, \texttt{AJCR937787} for a single-case article), and is treated as an
independent case vector downstream.

The final output is a directory of 391 per-article JSON files. Each file contains a
\texttt{metadata} block, a structured \texttt{abstract}, a \texttt{cases} list with one
entry per patient holding the isolated \texttt{main\_text} and the patient's associated
figures and tables, and a \texttt{general} block for unclaimed assets. A processing
summary CSV records parsed status and per-article case count across all 392 entries.

\vspace{-5pt}

\subsubsection{Output Structure}

The final output is a dataset of JSON files where each entry contains a global \texttt{metadata} object, a structured \texttt{abstract}, and a list of \texttt{cases}. Each case object contains the isolated patient narrative (\texttt{main\_text}) and references to the specific data assets (tables/figures) relevant to that patient.
\vspace{-5pt}

\subsection{Anchor Resolution: \cui and ICD-10 Mapping}
\label{app:anchorresolution}

\vspace{-5pt}
Mapping raw clinical strings to canonical UMLS \cuis and ICD-10 codes proved non-trivial; we briefly document the failure of some approaches that motivated the eventual 2-stage pipeline design.
\vspace{-5pt}
\paragraph{LLMs and APIs as ontology resolvers: counterproductive}
Initial attempts to use MedGemma~\cite{sellergren2025medgemma} as a semantic resolver failed entirely; the model confidently hallucinated non-existent \cuis and fabricated terms. This is unsurprising, as LLMs encode token distributions rather than versioned database mappings. This aligns with prior findings that LLMs struggle with ICD-10 labeling (26\% accuracy)~\cite{Simmons2025Extracting} and exhibit high hallucination rates (35\%) even in specialized RAG setups~\cite{klang2024assessing}, making current LLMs highly unrealistic for reliable ontology resolution.

The UMLS REST API proved similarly unsuitable for large-scale annotation. Exact-string matching yielded poor recall, while fuzzy searches caused unacceptable precision losses by highly ranking superficial string overlaps. Since tuning search parameters per query is unscalable, we abandoned both approaches. Instead, we implemented a tried-and-true deterministic NER pipeline using \texttt{medspaCy} and \texttt{QuickUMLS} locally against a fixed UMLS release.
\vspace{-5pt}

\paragraph{Working solution: medspaCy $\rightarrow$ QuickUMLS.} 
In the first stage, clinical named entity recognition is performed using \texttt{medspaCy} \cite{medspacy}, a spaCy-based framework for clinical text processing, which segments input text and identifies candidate medical concept spans.  In the second stage, each candidate span is passed to \texttt{QuickUMLS}~\cite{soldaini2016quickumls}, a fast approximate string matching engine built over a local UMLS installation, which returns the best-matching \cui along with a similarity score retained as the \texttt{mapping\_confidence}.  All additional non-anchor patient information, across all three sources, is mapped back to either \cui identifiers or, where not possible, tightly structured enumeration options or numeric scale representations. 

\vspace{-5pt}
\subsection{PatientVector: Supplementary Details}
\label{app:patientvector}
\vspace{-5pt}

The structural overview of the \pv schema is provided in Section~\ref{sec:patientvector}. Below are specific technical implementations and schema attributes excluded from the main text for brevity:

\begin{itemize}[noitemsep,topsep=0pt]
    \item \textbf{GroundTruth (\texttt{provenance\_tier}):} In addition to the MIMIC-IV tiers, reliability labels include \texttt{ddxplus\_gold} (synthetic ground truth) and \texttt{ajcr\_gold} (peer-reviewed case report ground truth).
    \item \textbf{PatientFacts Extensions:} The \texttt{Demographic} block also captures education level. \texttt{FamilyHistory} is strictly structured as a dictionary mapping specific family members to their associated \cui-mapped diagnoses or symptoms.
    \item \textbf{Explicit Symptom Negation:} Within \texttt{CasePresentation}, symptoms carry an explicit negation flag. This structurally distinguishes symptoms the patient actively denies from symptoms that are simply not mentioned (\texttt{not\_known}). This distinction is critical for evaluating whether an examiner model makes unsafe closed-world assumptions during clinical reasoning.
\end{itemize}
\vspace{-5pt}

\vspace{-5pt}

\section{Utterance Generation Details}
\label{app:utterance_details}
\vspace{-5pt}

To generate natural language patient utterances from the structured symptom
representations in each case presentation variant, we make use of
\href{https://huggingface.co/microsoft/UserLM-8b}{UserLM-8B}~\cite{naous2025flipping}, a
model released by Microsoft Research that is trained to produce human-like user
utterances rather than assistant-style responses. Unlike standard instruction-tuned
models, UserLM-8B is optimised to express information the way a person would (colloquially, incompletely, and with the natural hedging and imprecision of lay
speech) making it well-suited for converting structured clinical evidence entries
into plausible patient utterances.

For each symptom or evidence entry in the visible presentation layer of a variant
$CP_{xi}V_j$, UserLM-8B is used to generate a set of utterance candidates spanning a
range of expression styles: from relatively precise phrasing (``I have been experiencing
blurred vision'') to lay and colloquial formulations (``I can't even read or write, my
vision is blurry''), and from direct self-report to third-party reporting (``my wife
says I seem confused''). The persona parameters of the variant condition the generation to ensure that utterances
are consistent with the patient profile. The result is a pool of grounded, \cui-anchored
utterance candidates per symptom per patient variant per \pv, which together constitute a structured collection of pre-generated patient expressions, reproducible and decoupled from live "utterance generation" inference.

\begin{figure*}[t]
  \includegraphics[width=0.65\linewidth]{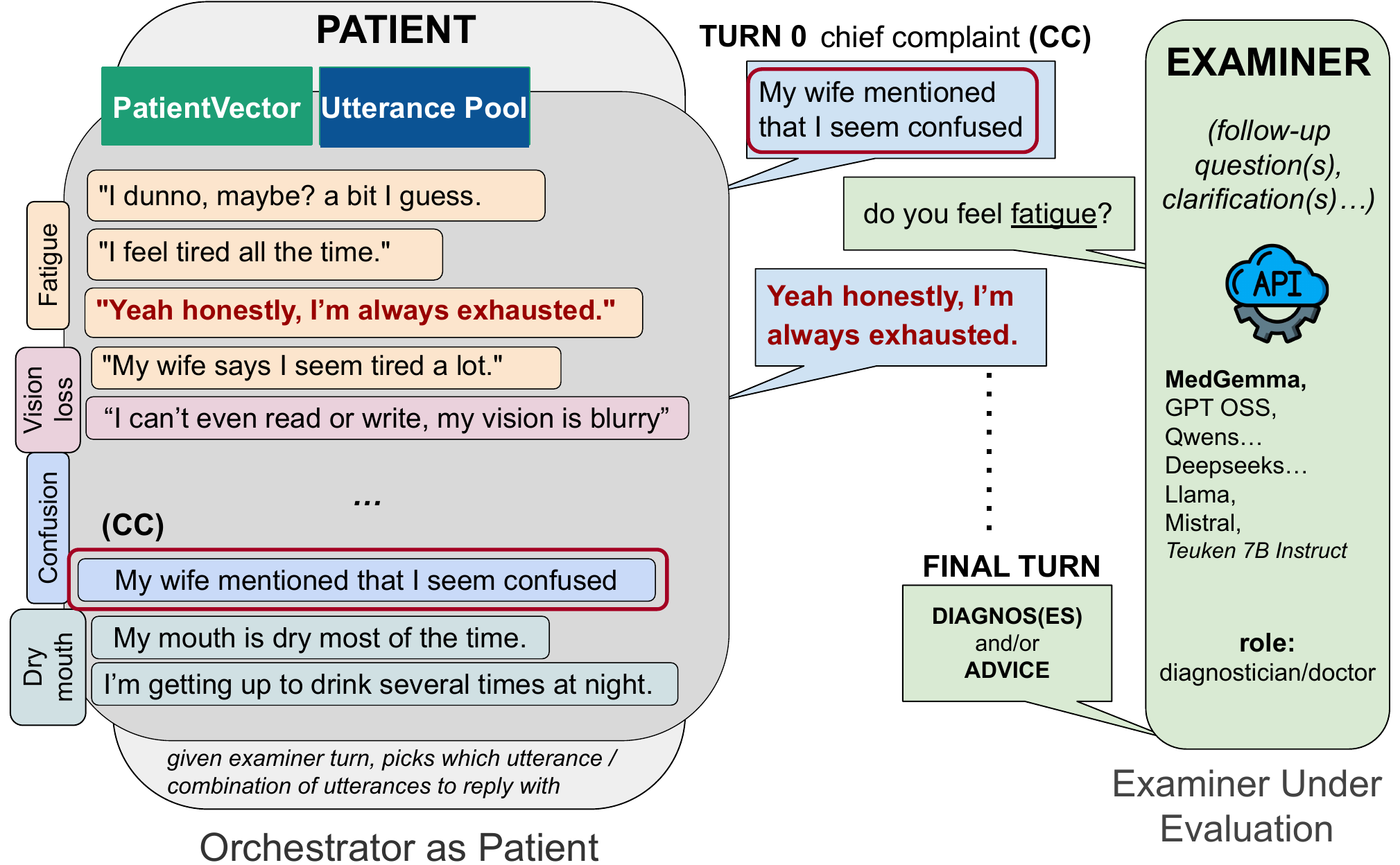}
    \centering
  \setlength{\belowcaptionskip}{-15pt}
  \caption{\textbf{Dialogue runtime}: The Patient Orchestrator receives the Examiner's turn and selects which pre-generated utterances to surface, drawing exclusively from the utterance pool for the current case. Content (what the patient says) is fixed at dataset creation time; selection logic (which utterance to return) is handled at runtime. The Examiner LM receives only the chief complaint at Turn~0 and issues follow-up questions until a terminal diagnosis or turn limit is reached.}
\label{fig:patientorchestrator}
\end{figure*}

\vspace{-5pt}

\section{Patient Orchestrator}
\label{app:patientorchestrator}
\vspace{-5pt}

At dialogue runtime, the Patient agent acts as an \textbf{orchestrator}: given the
Examiner's most recent question or clarification, it selects which pre-generated
utterance or combination of utterances best responds to that question, subject to the
reveal conditions defined in the variant config (illustrated in Fig.~\ref{fig:patientorchestrator}). This is not free generation, the
Patient draws from the pre-generated utterance pool for the current variant, ensuring
that responses remain grounded in the case facts, and are not subject to the "orchestrator" LLM generation. MedGemma~\cite{sellergren2025medgemma} is used
as the orchestration model, responsible for interpreting the Examiner's question in
context and selecting the appropriate patient response from the candidate set. This
approach deliberately separates the \textit{content} of what the patient says (grounded
in real clinical data, pre-generated) from the \textit{selection logic} (contextually
driven by the dialogue history), avoiding both the rigidity of purely rule-based
response selection and the uncontrolled hallucination risk of fully free-form generation by User-LM.

The Examiner agent receives only the chief complaint utterance at Turn 0 and issues follow-up
questions and clarification requests across subsequent turns. The
dialogue continues until the Examiner produces a terminal diagnosis or a maximum turn
limit is reached. Every turn role, raw text, Examiner question, Patient utterance
selected, and \cuis extracted is logged to a structured JSON log that serves as
the input to the evaluation pipeline.

\vspace{-5pt}

\section{Clinical Reasoning Metrics}
\label{app:metrics}
\vspace{-5pt}

The evaluation framework MTDiag enables goes beyond diagnostic accuracy. Because every symptom and diagnosis in the dataset is anchored to a \cui or ICD-10 code, evaluation can operate over ontological identifiers rather than surface strings, allowing graded, clinically interpretable scoring at both the dialogue and utterance level.  We describe the metrics below, organized from the most direct (diagnostic outcome) to the most structurally demanding (reasoning fidelity and harm).

\vspace{-5pt}

\subsection{Diagnostic Outcome Metrics}
\vspace{-5pt}

\paragraph{Diagnosis Accuracy ($\mathrm{Diag_{acc}}$).}
The trivial binary baseline: did the examiner model produce a diagnosis matching
the ground-truth ICD-10 code?

\begin{equation}
\mathrm{Diag_{acc}} =
\begin{cases}
1 & \text{if predicted code matches} \\
0 & \text{otherwise}
\end{cases}
\end{equation}

\vspace{-5pt}

\paragraph{Semantic Diagnostic Distance.}
Because ICD-10 is hierarchical, a binary match/miss discards clinically meaningful information.  A model outputting \texttt{I21.0} (acute myocardial infarction) when the ground truth is \texttt{I21.02} (STEMI of the LAD artery) has committed an under specification, not categorical error (difference in 3-character code). 
Secondary diagnoses can be used as a near-miss buffer: a predicted code matching any secondary diagnosis is flagged as clinically consistent rather than wrong.
\vspace{-5pt}

\subsection{Symptom Elicitation Metrics}
\vspace{-5pt}

The following metrics assess not whether the model reached the right diagnosis, but whether it asked the right questions to get there.  They require the \cui extraction step applied to examiner outputs at each turn~(Section~\ref{sec:orchestrator}).
\vspace{-5pt}

\paragraph{Disease-wise Symptom Score ($S_\mathrm{dise}$).}
Adapted from \citet{macherla2023mddial}, this metric evaluates how many of the symptoms the examiner queried are relevant to the diagnosed disease, adjusted by a turn-efficiency penalty.  Let $S_d$ be the set of symptoms associated with disease $d$ (drawn from the HPO
disease-phenotype prevalence table or curated lookup tables), and let $S_\mathrm{diag}$ be the \cui-resolved set of symptoms the examiner asked about.

\begin{equation}
f(s, S_d) = \mathbf{1}[s \in S_d]
\end{equation}

\begin{equation}
C = \frac{\min(N_\mathrm{gold},\, N_\mathrm{pred})}
         {\max(N_\mathrm{gold},\, N_\mathrm{pred})}
\end{equation}

\begin{equation}
S_\mathrm{dise} = \frac{C}{N_\mathrm{pred}}
  \sum_{s_i \in S_\mathrm{diag}} f(s_i,\, S_d)
\end{equation}

where $N_\mathrm{gold}$ is the number of turns in a reference "ideal" dialogue and $N_\mathrm{pred}$ is the examiner's turn count. The "ideal" dialogue is only well-defined for DDXPlus instances in MTDiag. For the other datasets, we can consider $N_\mathrm{gold}$ as the number of \cui-mapped symptoms in $S_d$ for the ground-truth diagnosis that is, the size of the relevant symptom set a thorough examiner would ideally cover and $N_\mathrm{pred}$ as the number of turns the examiner took. High $S_\mathrm{dise}$ indicates that the model's symptom inquiries were clinically coherent with the diagnosis it ultimately reached.
\vspace{-5pt}

\paragraph{MTDiag extension: weighted relevance}
The binary $f(s, S_d)$ treats all associated symptoms as equally important, which is clinically unrealistic.  We replace it with a frequency-weighted version drawn from the HPO \texttt{phenotype.hpoa} annotations:

\begin{equation}
f_w(s, S_d) = \omega(s, d) \cdot \mathbf{1}[s \in S_d]
\end{equation}

where $\omega(s, d) \in \{1.0, 0.75, 0.5, 0.25, 0.1\}$ (suggestion) corresponds to HPO frequency qualifiers \textit{obligate}, \textit{very frequent}, \textit{frequent}, \textit{occasional}, and \textit{very rare} respectively.  This rewards the prioritization of high-yield symptoms early in the dialogue.
\vspace{-5pt}

\paragraph{Symptom Precision and Recall}

\begin{equation}
\mathrm{Prec} = \frac{|S_\mathrm{diag} \cap S_d|}{|S_\mathrm{diag}|}
\end{equation}

\begin{equation}
\mathrm{Rec} = \frac{|S_\mathrm{diag} \cap S_d|}{|S_d|}
\end{equation}

Precision penalizes irrelevant or off-topic questioning; recall captures whether the model elicited a sufficient fraction of the relevant symptom space.
\vspace{-5pt}

\subsection{Reliability and Reasoning Fidelity}
\vspace{-5pt}

\paragraph{Reliability Score ($R_\mathrm{score}$).}
Also adapted from \citet{macherla2023mddial}, this metric unifies diagnostic and symptom performance: a model is credited only if it reached the correct diagnosis \textit{and} its symptom inquiries were sufficiently relevant.

\begin{equation}
R_\mathrm{score} =
\begin{cases}
1 & \text{if } S_\mathrm{dise} \geq t \text{ and } \mathrm{Diag_{acc}} = 1 \\
0 & \text{otherwise}
\end{cases}
\end{equation}

The threshold $t \in (0,1)$ controls strictness.  A model that guesses the correct diagnosis while asking clinically incoherent questions receives $R_\mathrm{score} = 0$, distinguishing systematic reasoning from lucky outcomes.

\paragraph{Pathognomonic and Sine Qua Non Errors.}
Some symptoms carry special logical status for a given diagnosis. \textit{Pathognomonic} symptoms are sufficient to confirm a diagnosis when present; \textit{sine qua non} symptoms are necessary: their absence invalidates the diagnosis.  A model that reaches a correct diagnosis without ever eliciting a sine qua non symptom has committed a logical failure: the conclusion is unsupported by the evidence gathered. Such cases can be flagged
\textbf{diagnostic hallucinations}: correct outputs reached via clinically incoherent pathways.

Pathognomonic errors measure whether the model failed to inquire about or recognize a uniquely identifying symptom for its own stated diagnosis.  Sine qua non errors capture omission of necessary symptoms. Both can be integrated into $R_\mathrm{score}$ as hard veto conditions: a diagnosis unsupported by its sine qua non evidence is treated as
invalid regardless of string match. 

\paragraph{Anchoring Bias}
A model may pursue a single working hypothesis throughout the dialogue, ignoring evidence that would refocus the differential. Using disease-phenotype associations, this can be detected as the degree to which the examiner's queried symptoms
$S_\mathrm{diag}$ are contained within $S_d$ for a single candidate diagnosis $d$, with no exploration of symptoms associated with plausible alternatives. A model that asks only cardiovascular-related questions while the case is a pulmonary embolism is anchoring.  Anchoring could lead to a correct diagnosis only when the diagnosis matches the model's opening hypothesis and constitutes an error in all other cases.

\paragraph{Premature Closure.}

If a model commits to a diagnosis before eliciting a sufficient proportion of the presenting symptoms by symptom-frequency weight, the diagnosis is flagged as premature.  No established clinical threshold exists for this criterion; it is left as a configurable parameter chosen to represent a minimal evidentiary bar before diagnostic commitment.Sensitivity analysis across different cutoff values is left to future work.

\paragraph{Turn Efficiency.}
The cost penalty $C$ in $S_\mathrm{dise}$ captures efficiency implicitly, but turn count can also be reported directly as a standalone metric.  Pathologically short dialogues (fewer turns than there are sine qua non symptoms) and pathologically long ones (exceeding the maximum turn limit) are flagged.

\end{document}